\documentclass[10pt,twocolumn,letterpaper]{article}

\usepackage[T1]{fontenc}
\usepackage[utf8]{inputenc}
\usepackage{libertinus} 
\usepackage[hyphens]{url}
\usepackage{microtype}
\usepackage{graphicx}
\usepackage{booktabs}
\usepackage{amsmath,amssymb}
\usepackage[round,authoryear]{natbib}
\usepackage[dvipsnames,table]{xcolor}
\usepackage[font=small,labelfont={bf,color=WSEBlue},labelsep=period]{caption}
\usepackage[letterpaper,top=0.66in,bottom=0.72in,left=0.72in,right=0.72in,
            columnsep=0.25in,headheight=32pt,headsep=0.12in]{geometry}
\usepackage{titlesec}
\usepackage{enumitem}
\usepackage{fancyhdr}
\usepackage{marvosym}
\usepackage{balance}
\usepackage{hyperref}

\definecolor{WSENavy}{HTML}{123D63}
\definecolor{WSEBlue}{HTML}{1B5B8F}
\definecolor{WSEAzure}{HTML}{2B73A0}
\definecolor{WSESky}{HTML}{4D88AA}
\definecolor{WSESlate}{HTML}{2D5874}
\definecolor{WSEPale}{HTML}{EEF4F8}

\hypersetup{
  colorlinks=true,
  linkcolor=WSEBlue,
  citecolor=WSEAzure,
  urlcolor=WSEBlue,
  pdfborder={0 0 0},
  pdftitle={When Stories Evolve: Benchmarking LLM Storytelling Across Agent Architectures in Open-Ended World Simulations},
  pdfauthor={Yuqi Chen, Sixuan Li, Yunfeng Cai, Xueai Li, Ka Man Yan, and Ying Li}
}

\setlist[itemize,enumerate]{nosep,leftmargin=*}

\titleformat{\section}
  {\large\bfseries\color{WSENavy}}{\thesection}{0.55em}{}
\titleformat{\subsection}
  {\normalsize\bfseries\color{WSEBlue}}{\thesubsection}{0.5em}{}
\titleformat{\subsubsection}
  {\normalsize\bfseries\color{WSEAzure}}{\thesubsubsection}{0.5em}{}
\titleformat{\paragraph}[runin]
  {\bfseries\color{WSESlate}}{}{0pt}{}[.]
\titlespacing*{\section}{0pt}{2.2ex plus .5ex}{1.2ex}
\titlespacing*{\subsection}{0pt}{1.8ex plus .4ex}{0.8ex}
\titlespacing*{\subsubsection}{0pt}{1.5ex plus .3ex}{0.6ex}
\titlespacing*{\paragraph}{0pt}{1.1ex}{0.65em}

\renewcommand{\headrulewidth}{0.3pt}
\renewcommand{\headrule}{%
  \raisebox{6pt}[0pt][0pt]{%
    \hbox to\headwidth{\color{WSENavy}\leaders\hrule height \headrulewidth\hfill}}}

\newcommand{\papertitle}{When Stories Evolve: Benchmarking LLM Storytelling Across Agent Architectures in Open-Ended World Simulations}
\newcommand{\equalglyph}{\textcolor{WSEAzure}{\(\star\)}}
\newcommand{\correspondglyph}{\textcolor{WSEBlue}{\Letter}}
\newcommand{\equalmark}{\textsuperscript{\equalglyph}}
\newcommand{\correspondmark}{\raisebox{0.8ex}{\scriptsize\correspondglyph}}
\newcommand{\institutionlogos}{%
  \makebox[\headwidth][c]{%
  \begin{tabular}{@{}c@{\hspace{0.20in}}c@{\hspace{0.20in}}c@{\hspace{0.20in}}c@{}}
  \raisebox{-0.5\height}{\includegraphics[height=0.31in]{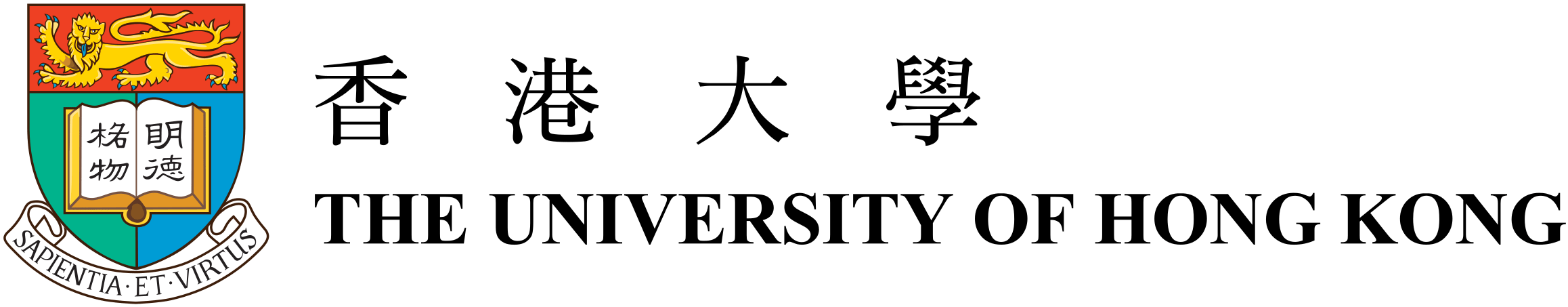}} &
  \raisebox{-0.5\height}{\includegraphics[height=0.27in]{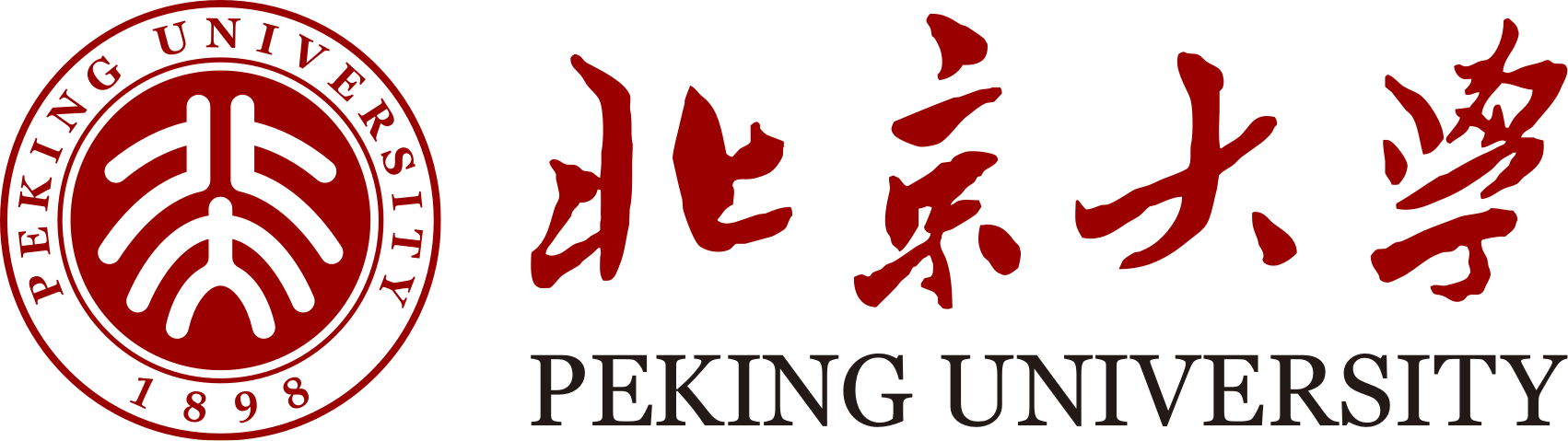}} &
  \raisebox{-0.5\height}{\includegraphics[height=0.27in]{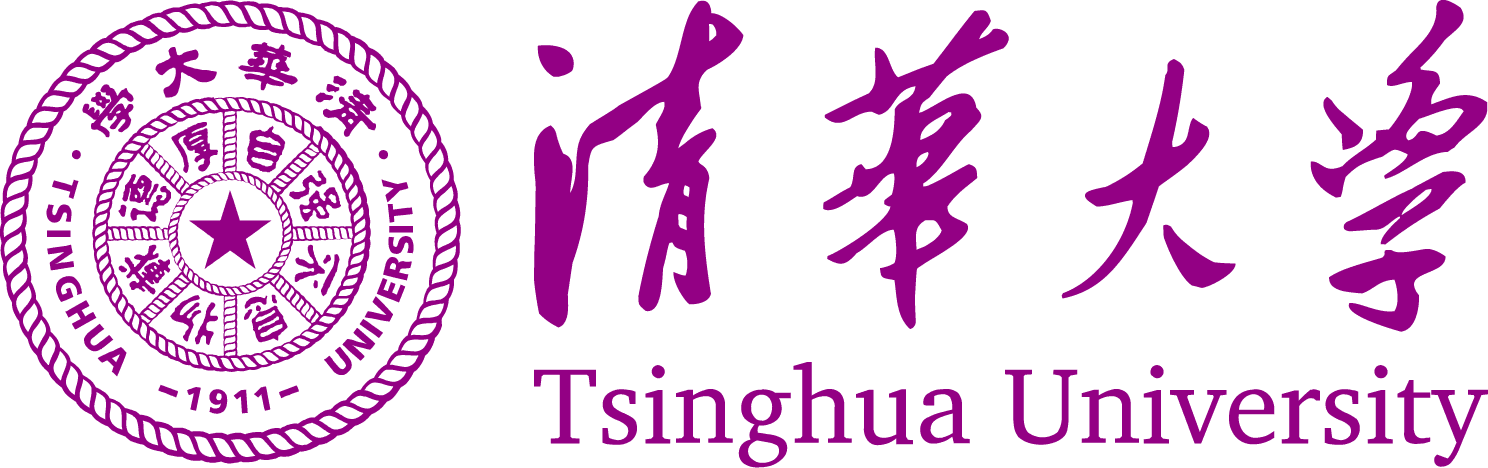}} &
  \raisebox{-0.5\height}{\includegraphics[height=0.27in]{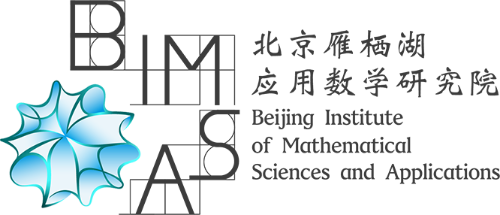}}
  \end{tabular}}}

\fancypagestyle{firstpage}{%
  \fancyhf{}%
  \fancyhead[C]{\institutionlogos}%
  \fancyfoot[R]{\footnotesize\color{WSESlate}\thepage}%
  \renewcommand{\headrulewidth}{0.3pt}%
  \renewcommand{\headrule}{%
    \hbox to\headwidth{\color{WSENavy}\leaders\hrule height \headrulewidth\hfill}}%
}

\newcommand{\WSEModelCount}{12}

\newcommand{\WSETrajectoryCount}{6,048}
\newcommand{\WSECellCount}{144}

\newcommand{\WSEITPConsistency}{82.1}
\newcommand{\WSEDraCorConsistency}{83.5}
\newcommand{\WSESmallvilleConsistency}{78.1}
\newcommand{\WSEDraCorMinusITP}{+1.4}
\newcommand{\WSEDraCorMinusITPLow}{-4.6}
\newcommand{\WSEDraCorMinusITPHigh}{+6.9}
\newcommand{\WSEDraCorMinusSmallville}{+5.4}
\newcommand{\WSEDraCorMinusSmallvilleLow}{-2.2}
\newcommand{\WSEDraCorMinusSmallvilleHigh}{+9.9}

\begin{document}
\thispagestyle{firstpage}

  \twocolumn[
\begin{@twocolumnfalse}
\begin{center}
{\fontsize{17.8}{21}\selectfont\bfseries\color{WSENavy}
\mbox{When Stories Evolve: Benchmarking LLM Storytelling}\\[-0.05em]
\mbox{Across Agent Architectures in Open-Ended World Simulations}\par}
\vspace{0.9em}
{\large
Yuqi Chen\textsuperscript{1}\equalmark\correspondmark
\quad Sixuan Li\textsuperscript{2}\equalmark
\quad Yunfeng Cai\textsuperscript{4}\\[0.35em]
Xueai Li\textsuperscript{1}
\quad Ka Man Yan\textsuperscript{1}
\quad Ying Li\textsuperscript{3,4}\correspondmark\par}
\vspace{0.65em}
{\small\color{WSESlate}
\textsuperscript{1}The University of Hong Kong
\quad \textsuperscript{2}Peking University
\quad \textsuperscript{3}Tsinghua University\\[-0.05em]
\textsuperscript{4}Beijing Institute of Mathematical Sciences and Applications (BIMSA)\par}
\vspace{0.55em}
{\small
\href{mailto:chenyuqi@hku.hk}{chenyuqi@hku.hk}
\quad
\href{mailto:liying0928@tsinghua.edu.cn}{liying0928@tsinghua.edu.cn}\par}
\vspace{0.55em}
{\footnotesize \equalglyph\ Equal contribution
\qquad \correspondglyph\ Corresponding authors\par}
\end{center}

\vspace{0.45em}
\noindent\colorbox{WSEPale}{%
\parbox{\dimexpr\textwidth-2\fboxsep\relax}{%
\vspace{0.35em}
\begin{abstract}
Large language models can write fluent stories, but open-ended storytelling requires more than local fluency. In evolving world simulations and AI-native games, models must preserve facts, relationships, causal dependencies, and character states as the world changes. We introduce WSE-bench, a process benchmark that separately evaluates sustained generation, canonical coherence, and meaningful development in dynamic LLM storytelling. Generation Coverage records the proportion of planned narrative steps produced; Consistency tracks when canon breaks; and Richness measures how meaningfully branching, player-shaped trajectories develop. Across frontier models, Consistency and Richness do not form a smooth trade-off: their empirical Pareto frontier is non-concave, with several non-dominated intermediate configurations that no positive linear weighting can select. Added structure can enrich trajectories, but it does not uniformly improve coherence and may shorten them. Model scale chiefly improves sustained generation, without producing reliable gains in canonical coherence or meaningful development. These results show that sustained generation, canonical coherence, and meaningful development are distinct and sometimes competing capacities. WSE-bench makes those dynamics visible by extending narrative evaluation from finished stories to the processes that create them.
\end{abstract}
\vspace{0.15em}}}
\vspace{1.2em}
\end{@twocolumnfalse}
]

\section{Introduction}

An open-ended story has no final draft. In evolving world simulations and LLM-driven open-narrative games, each action changes the state from which the next scene is generated \citep{park2023generative,han2024ibsen,wang2024storyverse}. A promise, injury, or transfer of knowledge can become binding evidence many steps later. Each continuation becomes part of the story's canon: the set of facts established so far. Open-ended storytelling can fail in three distinct ways: generation may stop, the story may contradict its accumulated canon, or it may remain coherent only by failing to develop. Evaluation must therefore distinguish sustained generation, canonical coherence, and meaningful development.

Most narrative evaluation focuses on completed stories, including book-length quality \citep{yang2025longstoryeval} and consistency bugs \citep{li2026lost}. Such settings do not reveal when an evolving trajectory first breaks canon, when development stalls, or whether an architecture trades one capability for the other. Interactive systems combine memory, planning, character, and world components \citep{park2023generative,han2024ibsen,chen2026storybox}, but differences in worlds, models, and player policies obscure component effects.

Long-horizon benchmarks expose compounding errors hidden by short evaluations \citep{ma2024agentboard,sinha2026diminishing}. Figure~\ref{fig:motivation} illustrates the three requirements at stake: sustained generation, canonical coherence, and meaningful development.

\newpage
\noindent\begin{minipage}{\columnwidth}
\centering
\includegraphics[width=0.98\columnwidth]{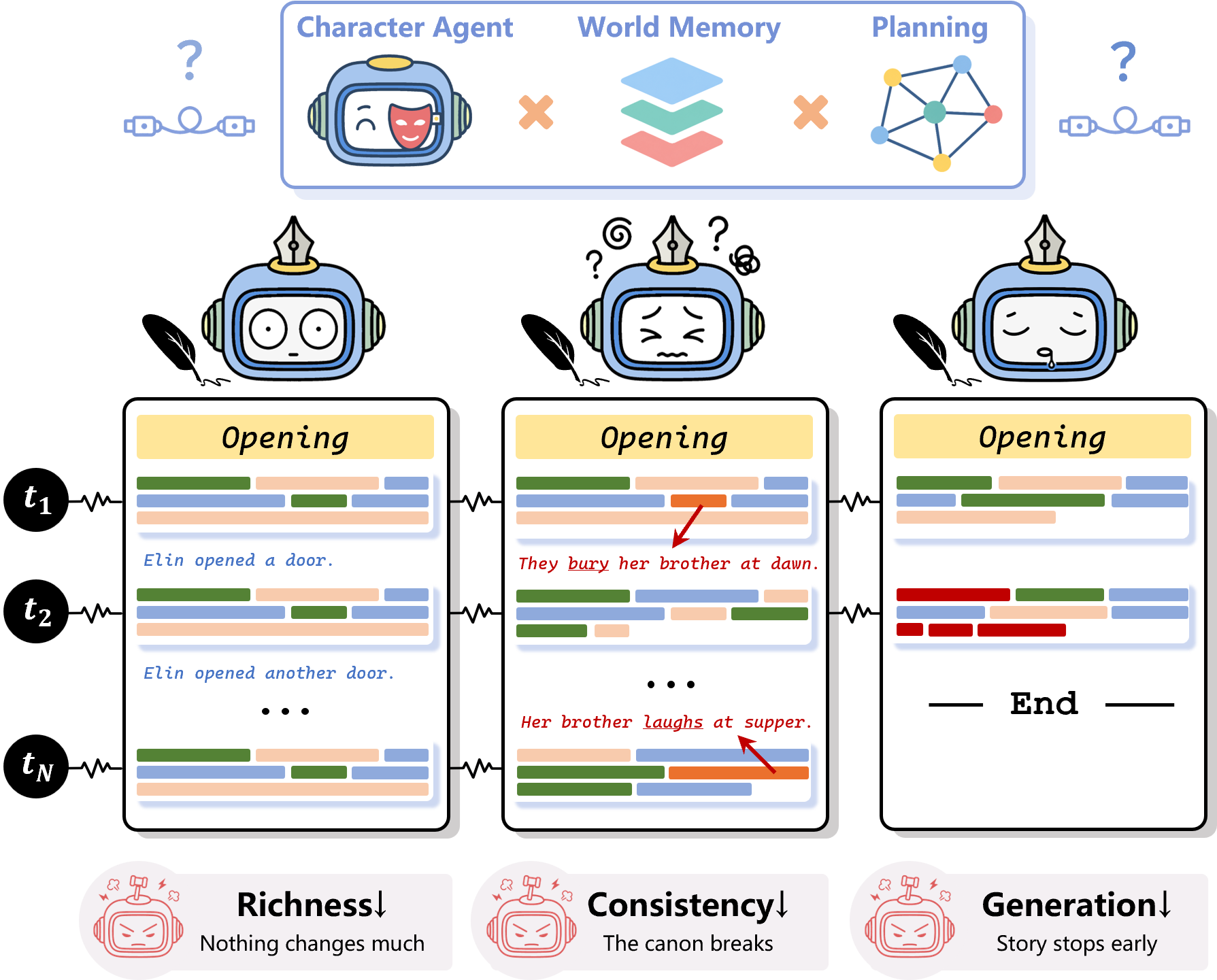}
\captionsetup{hypcap=false}
  \captionof{figure}{Conceptual requirements for open-ended storytelling across agent architectures: meaningful development (Richness), canonical coherence (Consistency), and sustained generation (Generation Coverage). WSE-bench evaluates evolving stories along these three dimensions.}
 \label{fig:motivation}
\end{minipage}
\par\vspace{0.7em}
To evaluate these requirements, we introduce \emph{WSE-bench} (\emph{When Stories Evolve Benchmark}), a process benchmark that operationalizes them as Generation Coverage, Consistency, and Richness. Generation Coverage records the proportion of planned narrative steps successfully produced. Each episode starts from a common initial world and a model-independent opening. A fixed player policy selects attempted actions, the tested architecture determines what follows, and independent evaluators identify contradictions and meaningful developments. Sharing the episode setup and choice-selection schedule across conditions makes architectural differences easier to interpret.

We examine how canonical coherence, meaningful development, and sustained generation vary with agent architecture and model choice, and whether dense-model scale produces consistent trends. Within-episode comparisons estimate architecture effects, while cross-model analyses characterize the models evaluated here.

Beyond architecture and model differences, we ask whether the origin of a world shapes the evolving canon. We hypothesize that source-derived worlds may be more vulnerable to contradiction when generated trajectories diverge from plots potentially seen during pretraining. Comparing simulation-native settings with worlds adapted from published plays tests this possibility, which we call \emph{prior-canon interference}.

Together, paired architecture comparisons, cross-model analyses, and blinded validation separate story quality from sustained generation. They distinguish configurations that expand the attainable Consistency--Richness region from those that move along its boundary.

\begin{figure*}[!t]
\centering
\includegraphics[width=\textwidth]{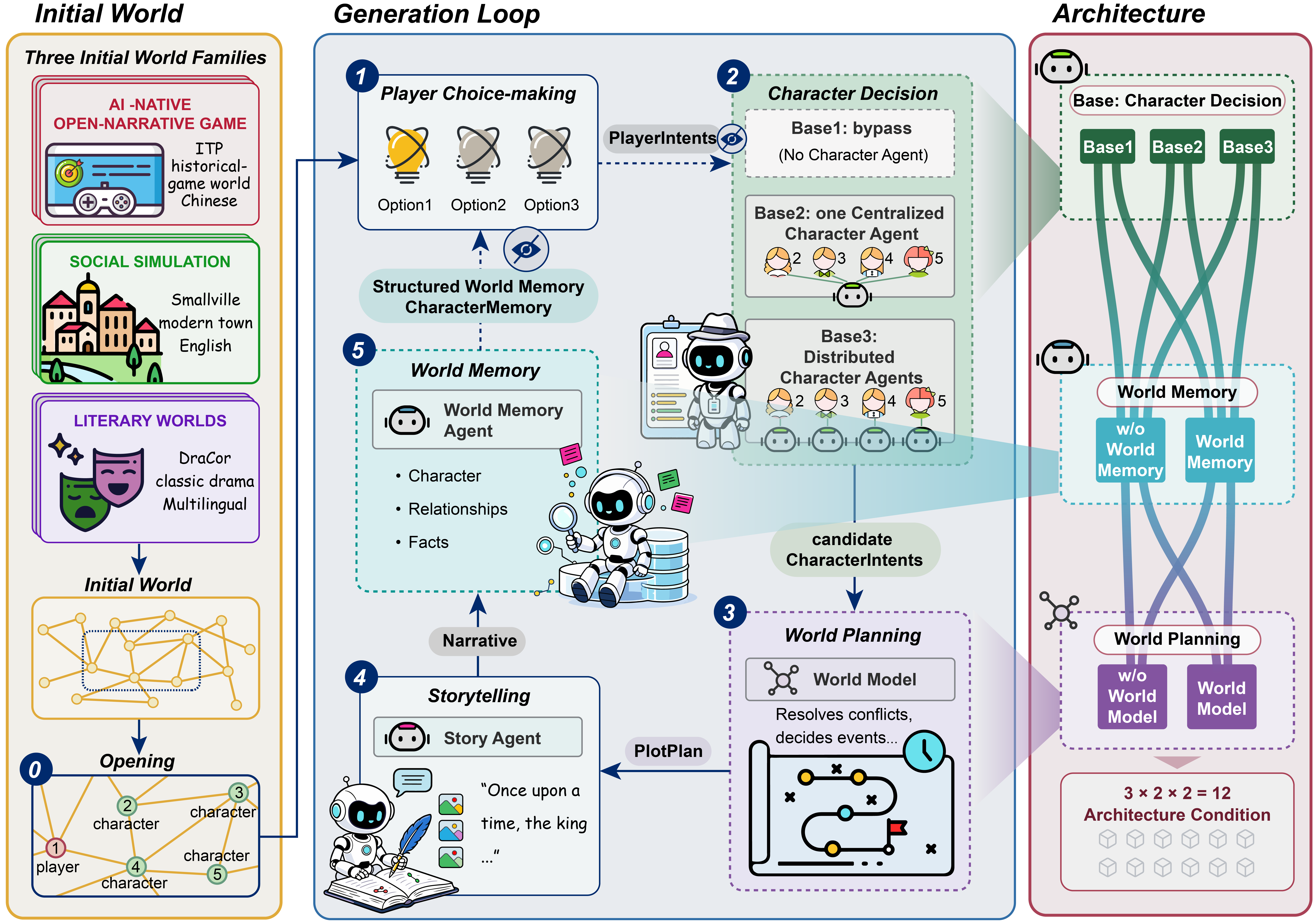}
\caption{WSE-bench combines standardized episode construction, recurrent generation loop, and systematic comparison of agent architectures. Each episode contains one player-controlled character and four non-player characters. At each step, \(B_1\) lets the Story Agent decide consequences directly, \(B_2\) uses one Character Agent to propose for all non-player characters, and \(B_3\) uses one isolated Character Agent per non-player character; the Story Agent remains the sole narrator. WMOD optionally organizes proposed actions into a pre-narration PlotPlan of at most 12 beats, whereas WMEM optionally records a structured post-narration world-state update. Crossing the three bases with the two binary modules yields 12 architecture conditions.}
\label{fig:overview}
\end{figure*}

\section{Related Work}

\paragraph{Planning and long-form story generation}
Planning-based generators separate storyline construction from surface realization \citep{yao2019planwrite}. Re3 conditions recursively on a global plan and current story state \citep{yang2022re3}. DOME combines dynamic outlines, temporal memory, and conflict analysis \citep{wang2025dome}, while FACTTRACK tracks temporally qualified world facts at the outline level \citep{lyu2025facttrack}. These systems establish planning and memory as plausible interventions, but generally compare bundled pipelines on completed drafts.

\paragraph{Narrative evaluation and LLM judges}
OpenMEVA tests story metrics for human agreement, discourse-level coherence, and robustness \citep{guan2021openmeva}. LongStoryEval organizes reader-relevant criteria for book-length narratives \citep{yang2025longstoryeval}, and ConStory-Bench targets consistency bugs in long stories \citep{li2026lost}. G-Eval motivates rubric-based LLM evaluation \citep{liu2023geval}, but creative-writing preference remains difficult even for strong judges \citep{fein2026litbench}. WSE-bench therefore uses dimension-specific Judges, blinded human validation, and step-level evidence. This design locates contradictions and substantive developments while keeping generation failures separate from Consistency and Richness.

\paragraph{Role-playing and simulated story worlds}
Generative Agents and SOTOPIA study memory, planning, and multi-agent interaction \citep{park2023generative,zhou2024sotopia}. CharacterBox evaluates role-playing through character trajectories in a narrator-coordinated sandbox \citep{wang2025characterbox}. Other interactive systems combine authorial and character processes through director--actor control, narrative planning, or character simulation \citep{han2024ibsen,wang2024storyverse,chen2026storybox,yu2025multiagentstory}. BOOKWORLD and EvolvingWorld extend these ideas to persistent literary societies and co-evolving character--world state \citep{ran2025bookworld,zong2026evolvingworld}. These studies usually assess bundled systems.

\paragraph{Long-horizon and stateful agent evaluation}
AgentBoard evaluates multi-turn agents through process-level progress rather than final success alone \citep{ma2024agentboard}. Long-horizon studies also show how small per-step errors compound with task length \citep{sinha2026diminishing}. Language-model world simulators expose unreliable state transitions \citep{wang2024worldsimulators}, while trained text-based world models can learn coherent dynamics when behavioral coverage is adequate \citep{li2026wordworld}.

Across these lines of work, system comparisons often conflate differences in models, worlds, player behavior, and architecture. WSE-bench holds the world, choice-selection policy, narrator role, output format, and tested model fixed within paired episodes while varying character-decision structure, WMOD, and WMEM independently. It evaluates inference-time behavior through canonical coherence, meaningful development, and sustained generation over long-horizon, open-ended trajectories.

\section{Benchmark Framework}

WSE-bench represents each evolving story as a trajectory grounded in a common initial world. Figure~\ref{fig:overview} traces the framework from standardized episode construction through recurrent story generation to evaluation.  Operationally, \(B_1\) assigns character decisions to the Story Agent, \(B_2\) adds one Character Agent for all non-player characters, and \(B_3\) assigns one isolated Character Agent to each non-player character. WMOD supplies a pre-narration PlotPlan of at most 12 beats, while WMEM applies a structured state update after narration.  The resulting trajectories are evaluated with Generation Coverage, Consistency, and Richness.

\subsection{Initial Worlds and Episode Construction}

Each episode is grounded in an \texttt{InitialWorld}, a self-contained canonical starting state that specifies characters, relationships, persistent facts, and time. The benchmark collection used here draws from three source families. Two are simulation-native: Into-the-Painting (ITP), an open-narrative historical game world \citep{fish2026itp}, and Smallville, a simulated society \citep{park2023generative}. The third consists of literary worlds adapted from the multilingual DraCor drama collection \citep{fischer2019dracor}. The simulation-native settings do not prescribe a reference continuation. DraCor worlds, by contrast, begin from published plays whose plots and character arcs may be represented in model pretraining.

A common \texttt{InitialWorld} format makes architectures comparable across sources. Each \texttt{EpisodeSeed} defines a connected cast, the focal player, their initial relationships and locations, and the instance language. The collection contains 42 five-character seeds, 14 per source family, balanced for focal-player representation and relationship structure.

Each seed has a native-language opening created without an evaluated model. All 12 architecture conditions follow the same pre-specified choice schedule. The instances span Chinese, English, Dano-Norwegian, Russian, and Swedish. Appendix~\ref{sec:episode-construction} provides further details on episode construction and cross-architecture comparison.

\subsection{Evolving Trajectories}

The opening introduces the player-controlled character and multiple active non-player characters. At step \(t\), a deterministic policy selects an option from the previous narrative. The selected option expresses what the player attempts, and the agent system determines its consequences and continues the story.

Each step produces a narrative continuation \(N_t\), options for the next action, and indicators of whether the story advances to a new day or ends. To keep the source of canon unambiguous, only the generated narrative can establish new story facts. The history supplied to later steps contains selected player intents, generated narratives, and time. When WMEM is enabled, it records a structured state update after each narrative, but that update cannot rewrite canon. A failed update therefore does not invalidate an already generated step.

This trajectory-centered view does not prescribe a correct continuation. Instead, each new step is evaluated against the canon accumulated along its own path and by how meaningfully it develops that path.

\subsection{Metrics}

Generation Coverage measures how much of the planned trajectory is produced; Consistency and Richness measure its canonical coherence and meaningful development.

\subsubsection{Generation: Trajectory Coverage}

For trajectory \(i\), let \(L_i\) be the number of valid narrative steps available for evaluation, and let \(H\) be the planned horizon. For a cell containing \(n\) trajectories, Generation Coverage is
\begin{equation}
G=\frac{100}{nH}\sum_{i=1}^{n}L_i.
\end{equation}
Trajectories with no valid narrative have \(L_i=0\). Thus, \(G\) is the mean percentage of planned narrative steps observed across trajectories. It measures the ability to sustain generation through the planned horizon, not narrative quality. Continuations after generation failures or valid story endings remain unobserved.

\subsubsection{Consistency: Canonical Coherence}
A narrative step is inconsistent only when all of the following conditions hold:
\begin{enumerate}
    \item the starting state or an earlier narrative explicitly establishes a fact, or unavoidably entails it;
    \item the current narrative explicitly establishes or unavoidably entails a conflicting claim about the same entity, reference, and applicable time; and
    \item the current narrative contains no explicit or clearly implied transition that reconciles the two.
\end{enumerate}
These criteria distinguish contradiction from narrated change. Missing information and unfulfilled player intents are not inconsistencies; style, plausibility, repetition, and progress are evaluated separately. The first step meeting these criteria marks the onset of inconsistency; uncertain cases remain separate.

A trajectory that ends because generation fails has not demonstrated consistency over the missing future steps, but neither has it demonstrated a contradiction. Such trajectories are therefore right-censored, and a Kaplan--Meier curve estimates the proportion remaining contradiction-free at each step \citep{kaplan1958nonparametric}. Let \(\widehat S_C(t)\) denote this estimate, and let \(Y_t\) be the number of trajectories still observed at step \(t\) with no earlier contradiction. The two Consistency scores are
\begin{equation}
\begin{aligned}
C_{\mathrm{obs}}
&=\frac{100}{H}\sum_{t=1}^{H}\widehat S_C(t),\\
C_{\mathrm{adj}}
&=\frac{100}{H}\sum_{t=1}^{H}\mathbb{I}[Y_t>0]\widehat S_C(t).
\end{aligned}
\end{equation}
The primary score, \(C_{\mathrm{obs}}\), carries forward the last estimable value if no trajectory remains under observation. The conservative alternative, \(C_{\mathrm{adj}}\), instead assigns zero after observational support is exhausted. This prevents a low-coverage cell from appearing artificially favorable in the Pareto comparison without treating generation failure as a contradiction or imputing unseen content. Appendix~\ref{sec:two-consistency-estimands} gives the Kaplan--Meier and risk-set definitions.

\subsubsection{Richness: Meaningful Development}
Richness Judges identify text-supported units that introduce or advance consequential events, character or relationship changes, story threads, or world elements. Repetition, stylistic elaboration, and text length alone do not count.

The Richness score \(R\) combines the frequency and breadth of meaningful development. The three breadth dimensions capture event diversity, meaningful character and relationship change, and consequential world or social expansion. Let \(Q\in[0,1]\) be the log-scaled frequency of non-repetitive development per observed narrative step, capped at the 95th percentile of the calibration data. Let \(B\in[0,1]\) be the mean normalized rating across these dimensions. Then
\begin{equation}
R=100\sqrt{QB}.
\end{equation}
The geometric mean prevents abundant but narrow activity from fully compensating for broad but sparse development. Appendix~\ref{sec:richness-computation} gives the exact calculation of \(Q\) and \(B\).

Consistency and Richness remain independent during evaluation. A canon-contradicting development still counts toward Richness and is penalized only by Consistency. Conversely, a static trajectory can remain coherent while receiving low Richness. Because these qualities can diverge, the results compare them jointly without collapsing them into a single score.

\section{Agent Architectures}

The evaluated grid combines one of three base architectures with zero, one, or two add-on modules:
\begin{equation}
    \begin{aligned}
    \mathcal{A}_{b,w,p}
      &=B_b\oplus \mathrm{WMEM}_w\oplus \mathrm{WMOD}_p,\\
    b&\in\{1,2,3\},\qquad w,p\in\{\mathrm{off},\mathrm{on}\}.
    \end{aligned}
\end{equation}
Here \(\oplus\) denotes architectural composition, and \(w\) and \(p\) indicate whether WMEM and WMOD are enabled. Each base therefore has four module configurations: neither module, memory only, planning only, or both. For each tested model, the same model and operating mode are used for every enabled component.

\(B_1\) has no Character Agent. The shared Story Agent decides consequences and narrates. \(B_2\) uses one Character Agent to propose actions jointly for all non-player characters, whereas \(B_3\) uses a separate Character Agent for each non-player character. These bases reflect three character-action topologies from prior work: single-narrator long-form generation (\(B_1\)) \citep{yang2022re3}, centralized multi-character coordination (\(B_2\)) \citep{zong2026evolvingworld}, and decentralized character agents (\(B_3\)) \citep{park2023generative,han2024ibsen,yu2025multiagentstory}.

WMOD inserts a stateless World Model before narration. It adjudicates the selected player attempt together with any Character-Agent proposals and returns a PlotPlan of at most 12 dependency-linked beats; the Story Agent receives this plan in place of the raw proposals. WMEM runs after narration: a World Memory Agent compares the realized narrative with prior structured state, emits a MemoryDelta of persistent changes, and a rule-based reducer updates character conditions, relationships, facts, and supporting characters for the next step. The Story Agent remains the sole narrator, and only the initial world and generated narratives establish canon. Paired conditions share the episode setup, choice-selection schedule, Story Agent role, and tested model. Appendix~\ref{sec:episode-construction} shows the full information flow, and Appendix~\ref{sec:architecture-contrasts} gives results for each base architecture.

 \section{Evaluation and Validation}

\subsection{LLM-Judge Evaluation}

Evaluators receive the \texttt{InitialWorld}, opening, and generated trajectory, but no model, architecture, internal generation data, or source-revealing identifiers. Appendix~\ref{sec:evaluation-definitions} details these evaluation inputs.

Consistency Judges apply the three criteria above to every narrative step. To mark a contradiction, a Judge must identify the prior fact, the conflicting claim, their mutual exclusion, and the missing transition. Each Judge verdict has three states---consistent, inconsistent, or uncertain---and the uncertain state is retained through aggregation. Only a valid majority of consistent or inconsistent votes yields a binary step verdict for the first-contradiction risk set; cases with a missing Judge or no valid majority are excluded from that binary risk set. Richness Judges return the development units and dimension scores defined above , and  the trajectory-level median determines \(R\).

\subsection{Human Validation of Judges}
\label{sec:human-validation}

Human annotations   provide references on separate  held-out   samples, using the same rubrics without  model, architecture, or Judge outputs.   For Consistency, design-weighted  comparison of a valid  two-of-three   Judge majority with human labels on 120 trajectories reaches  85.7\% accuracy (95\% CI \(81.3\)--\(90.1\%\)) and 82.0\% balanced accuracy.   For Richness, the three-Judge median   on a separate 36-trajectory sample achieves  MAE \(7.63\) (95\% CI \(6.26\)--\(8.95\)) and Pearson \(r=.861\)  against the human reference. We use these validation results to assess the benchmark's aggregate Judge measures; Appendix~\ref{sec:evaluation-validation}   details sampling and agreement.

 {
 \subsection{Experimental Design and Analysis}

The evaluation   crosses  \WSEModelCount{} models,  42   shared episode seeds, and 12   architectures, yielding  \WSETrajectoryCount{} trajectories and \WSECellCount{}   cells. One model--architecture cell contains all 42 seeds; its  model fills every enabled LLM role. The Qwen   panel contains  dense Qwen2.5-Instruct   3B, 7B, 14B, 32B, and 72B checkpoints \citep{qwen25techreport}, plus  Qwen3.5 -397B-A17B MoE. The Frontier panel contains Grok-4.3, GLM-5.2, DeepSeek-V4-Pro, Kimi-K2.6, Gemini-3.1-Pro, and GPT-5.6-sol.  Panel summaries refer to and equally weight these twelve tested models.

 Architecture contrasts pair cells over shared seeds and weight evaluated models equally. Their intervals hierarchically bootstrap source clusters and \texttt{EpisodeSeed} records (Appendix~\ref{sec:architecture-contrasts}); source-family contrasts also weight models equally (Appendix~\ref{sec:source-family-analysis}). The heatmap reports \(C_{\mathrm{obs}}\) beside Generation Coverage, whereas Pareto analysis uses \(C_{\mathrm{adj}}\) to avoid carrying survival beyond observational support. Dense-Qwen, source-family, and horizon analyses respectively describe five-checkpoint association, the studied worlds, and when step-20 events first appear.  All analyses retain Generation Coverage, Consistency, and Richness as separate outcomes.

}
\section{Experiments and Results}

The experiments compare outcome profiles across models and architectures, then estimate paired architectural shifts, dense-Qwen scale associations, source-family patterns, and long-horizon failures. Qwen3.5 MoE is analyzed separately from dense-model scaling.

\begin{figure}[!t]
\centering
\includegraphics[width=0.97\columnwidth]{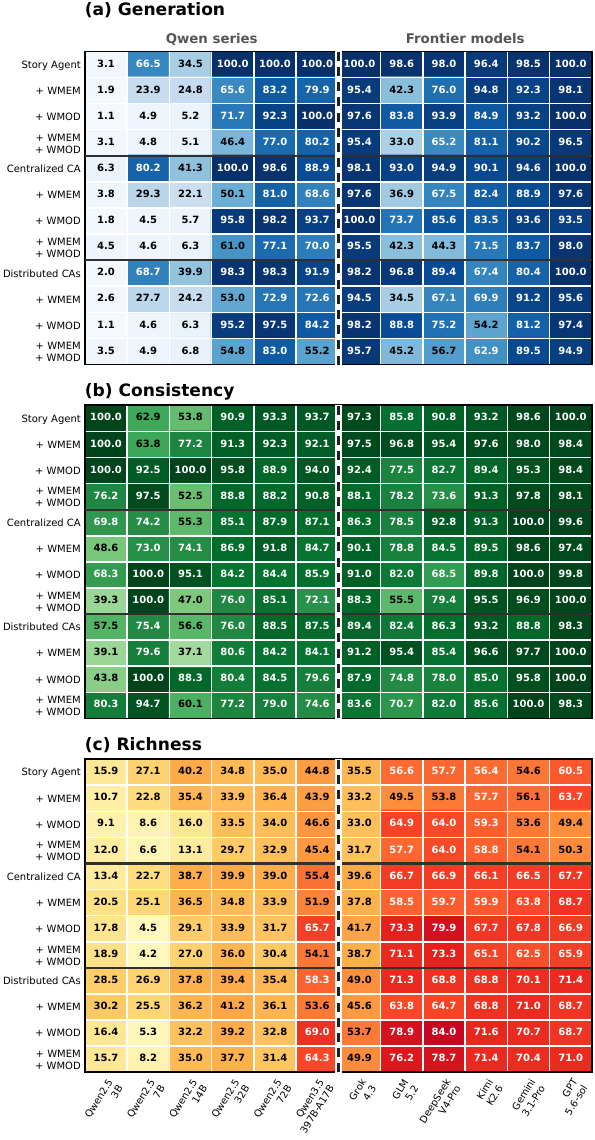}
\caption{Model-by-architecture point estimates for the Qwen and Frontier panels. Columns index models, and rows index the 12 architecture conditions. The three panels show Generation Coverage \(G\), Consistency \(C_{\mathrm{obs}}\), and Richness \(R\); the dashed rule separates the two analysis groups. Paired architecture contrasts below provide uncertainty intervals.}
\label{fig:model-architecture}
\end{figure}

Figure~\ref{fig:model-architecture} maps how Generation Coverage, Consistency, and Richness vary across the evaluated model--architecture conditions. Generation Coverage shows the clearest between-model separation, including an apparent gradient across the dense Qwen2.5 checkpoints. Consistency remains high for many Frontier conditions but changes heterogeneously across architecture rows, whereas Richness is generally greater in the \(B_2\) and \(B_3\) rows and produces an ordering distinct from Coverage and Consistency. The cross-panel outcome differences motivate Figure~\ref{fig:formal-pareto}'s joint Consistency--Richness analysis; the heterogeneous within-model architecture shifts motivate the paired contrasts in Table~\ref{tab:component-effects}; and the model and scale analysis below quantifies the apparent between-model ordering and checkpoint gradient.

\subsection{Consistency--Richness Trade-off}
\label{sec:consistency-richness-tradeoff}

\begin{figure}[!t]
\centering
\includegraphics[width=0.98\columnwidth]{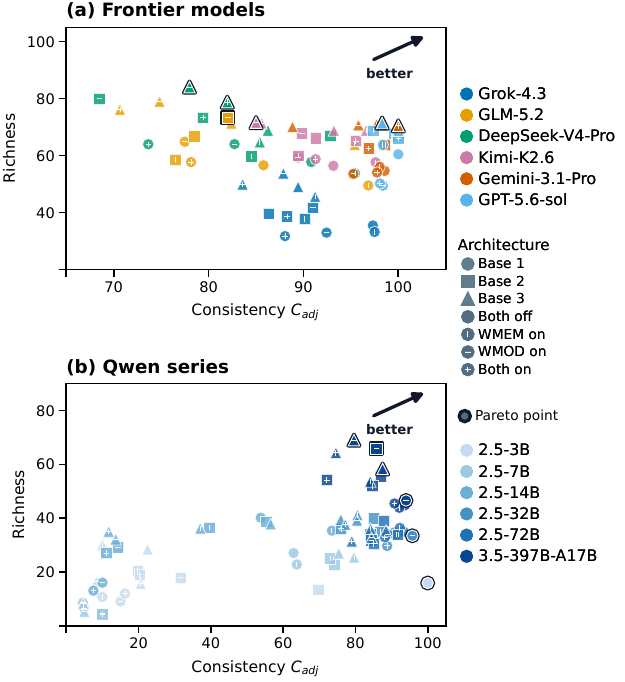}
\caption{Consistency--Richness point estimates based on \(C_{\mathrm{adj}}\), which sets the survival contribution to zero after a cell loses observational support. Shapes denote \(B_1\) (Story Agent only), \(B_2\) (one centralized Character Agent), and \(B_3\) (distributed Character Agents); white overlays mark WMEM, WMOD, or both. The upper-right boundary identifies non-dominated configurations, and supported configurations lie on its upper concave envelope.}
\label{fig:formal-pareto}
\end{figure}

At the plotted point estimates, Figure~\ref{fig:formal-pareto} reveals a non-concave empirical Consistency--Richness frontier: six model--architecture cells are non-dominated, but only two lie on the upper concave envelope. We use \(C_{\mathrm{adj}}\) so the comparison reflects contradiction-free survival while it remains observationally supported. Replacing \(C_{\mathrm{adj}}\) with \(C_{\mathrm{obs}}\) leaves the same pooled six-cell non-dominated set across all 144 cells (Appendix Figure~\ref{fig:supp-consistency-estimands}).

DeepSeek-V4-Pro and GLM-5.2 occupy the Richness-favoring region, GPT-5.6-sol and Gemini-3.1-Pro occupy the Consistency-favoring end, and Kimi-K2.6 lies between them. DeepSeek-V4-Pro B3+WMOD and Gemini-3.1-Pro B3+WMEM+WMOD are the two supported configurations. The four intermediate cells are \emph{unsupported}: they are non-dominated among observed configurations but cannot maximize any positive linear weighting of the two outcomes \citep{ehrgott2005multicriteria}. Every Grok-4.3 condition is dominated. The adjacent GPT-5.6-sol cell lies 0.05 Richness points below the envelope; this gap is small relative to sampling uncertainty, leaving that cell's supportedness uncertain.

All pooled-frontier cells come from Frontier models, while Qwen cells exhibit internal trade-offs. The four unsupported cells represent distinct observed compromises that a single weighted score would skip, showing why separate Consistency and Richness results preserve choices hidden by scalar ranking.

\subsection{Architectures, Models, and Worlds}

Beyond the joint Consistency--Richness geometry, the individual outcomes vary with architecture, model identity, dense-model scale, and \texttt{InitialWorld} source family. Architecture comparisons are paired within episodes.

\subsubsection{Agent Architecture Effects}

 \begin{table*}[!t]
\centering
\small
\setlength{\tabcolsep}{4pt}
\begin{tabular}{@{}lrrrr@{}}
\toprule
Contrast & $\Delta G$ & $\Delta C_{obs}$ & $\Delta C_{adj}$ & $\Delta R$ \\
\midrule
\multicolumn{5}{l}{\textit{Qwen series}} \\
B2 -- B1 & +0.8 [-1.0, +2.5] & -9.2 [-14.5, -3.2] & -3.4 [-8.3, +3.3] & +4.0 [+2.6, +6.0] \\
B3 -- B1 & -1.1 [-3.0, +0.9] & -12.0 [-17.7, -5.4] & -8.4 [-10.5, -1.7] & +7.0 [+4.4, +9.0] \\
B3 -- B2 & -1.9 [-3.5, -0.3] & -2.8 [-8.5, +3.6] & -5.0 [-7.7, +0.4] & +3.0 [+0.2, +4.8] \\
WMEM on -- off & -18.0 [-19.8, -16.1] & -5.6 [-10.2, -0.8] & -6.0 [-8.8, -0.6] & -1.3 [-2.5, +0.1] \\
WMOD on -- off & -13.7 [-15.6, -11.7] & +4.7 [-1.0, +11.9] & -22.9 [-25.7, -16.0] & -5.9 [-7.2, -4.2] \\
\addlinespace[1pt]
\multicolumn{5}{l}{\textit{Frontier models}} \\
B2 -- B1 & -4.1 [-6.2, -1.9] & -3.2 [-5.2, -1.2] & -3.3 [-5.3, -1.4] & +9.2 [+7.8, +10.5] \\
B3 -- B1 & -7.5 [-10.8, -4.1] & -2.7 [-5.2, -0.4] & -2.7 [-5.4, -0.4] & +14.6 [+13.3, +16.0] \\
B3 -- B2 & -3.4 [-6.1, -0.7] & +0.5 [-1.8, +2.8] & +0.6 [-1.9, +2.9] & +5.5 [+4.5, +6.5] \\
WMEM on -- off & -14.1 [-16.3, -11.9] & +0.3 [-1.3, +2.0] & +0.2 [-1.8, +1.7] & -2.4 [-3.0, -1.8] \\
WMOD on -- off & -5.5 [-7.2, -3.8] & -5.3 [-7.1, -3.4] & -5.2 [-6.9, -3.1] & +3.4 [+2.4, +4.2] \\
\bottomrule
\end{tabular}
\caption{Architecture effects on the four reported outcomes. Each contrast averages over the other architecture factors. Values are mean differences across the evaluated models, weighted equally, with 95\% hierarchical cluster-bootstrap intervals.}
\label{tab:component-effects}
\end{table*}

  Table~\ref{tab:component-effects} shows that paired changes to  character decision structure mainly   shift  the Consistency--Richness trade-off. Among Frontier models, B2 and B3 raise Richness by 9.2 and 14.6 points relative to B1  while reducing \(C_{\mathrm{obs}}\) by 3.2 and 2.7 points. B3 's independent proposals may increase both development and coordination burden, echoing  evidence that multi-agent collaboration can become counterproductive   under excess coordination demands  \citep{kim2026collaboration,sun2026perspectivegap}.

The effect of WMOD depends on the character-decision topology. For Frontier models, it changes Richness by only \(+0.4\;[-1.3,+2.1]\) points at B1. The gains are larger at B2 and B3: \(+4.3\;[+2.6,+6.0]\) and \(+5.3\;[+4.0,+6.6]\), respectively. This pattern is consistent with planning being most useful when the Story Agent must coordinate separately proposed character actions, although WMOD does not produce a corresponding Consistency gain. Appendix~\ref{sec:architecture-contrasts} gives the results for each base architecture.

The marginal effect of WMOD also differs across model groups. It raises Frontier Richness by 3.4 points but lowers Qwen Richness by 5.9 points, while reducing Generation Coverage in both groups. The same planning scaffold is therefore not uniformly useful; its effect depends on the model used with it.

WMEM provides no stable Consistency benefit at any base and reduces Generation Coverage in both groups. Because every condition retains the complete narrative history, this result concerns structured state added to full-history access, not memory as a substitute for history. One possible explanation is that long contexts can obscure distant evidence, while extracted summaries may lose information and introduce an additional failure point \citep{wu2025longmemeval,pollertlam2026beyond}. Stronger graph-memory systems often add retrieval, temporal provenance, and conflict handling \citep{banerjee2026apexmem,hu2026memorygraphs}. Appendix~\ref{sec:architecture-contrasts} gives model-specific results and module interactions.

\subsubsection{Model and Scale Effects}

When scores are averaged across architectures, Figure~\ref{fig:model-architecture} shows that no Frontier model leads all three outcomes. GPT-5.6-sol combines the highest mean Consistency with near-complete Generation Coverage (\(C_{\mathrm{obs}}=99.0\), \(G=97.6\)). DeepSeek-V4-Pro instead has the highest mean Richness (\(R=68.0\)) but lower Consistency and Generation Coverage. Grok-4.3 nearly matches GPT-5.6-sol in Generation Coverage (\(G=97.2\)) while producing substantially lower Richness (\(R=40.8\)). These profiles complement the Pareto analysis in Section~\ref{sec:consistency-richness-tradeoff} and show that model differences cannot be reduced to a single quality ordering.

Scale is a separate question that can be studied within a comparable model family. Across five dense Qwen2.5 checkpoints, Generation Coverage rises by an estimated 19.32 percentage points per parameter doubling. Appendix~\ref{sec:model-scale-analysis} provides the full analysis.

Larger parameter count does not yield a reliable monotonic improvement in either story measure. Consistency rises across much of the dense series, but low Generation Coverage among smaller models leaves unstable survival tails. Richness is also non-monotonic. Thus, within the tested Qwen2.5 range, parameter count strongly predicts sustained generation but not canonical coherence or meaningful development.

\subsubsection{Source-Family Effects}

Source-family comparisons test the prior-canon interference hypothesis: DraCor worlds may become more inconsistent than the simulation-native ITP and Smallville worlds after branching from potentially familiar plots. Each of the \WSEModelCount{} models receives equal weight.

\begin{figure}[!t]
\centering
\includegraphics[width=0.99\columnwidth]{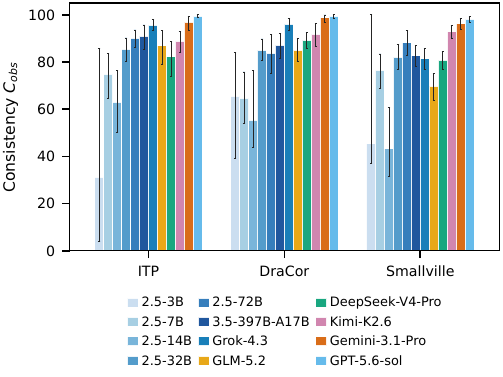}
\caption{Per-model Consistency \(C_{\mathrm{obs}}\) across the three world families, with each model pooled over 12 architectures. Markers and whiskers show model-specific means and 95\% hierarchical cluster-bootstrap intervals. The plotted profiles show within-model source-family variation; aggregate source-family contrasts give each of the 12 evaluated models equal weight.}
\label{fig:dracor-literary}
\end{figure}

Contrary to this hypothesis, Figure~\ref{fig:dracor-literary} shows that, across the 12 evaluated models, \(C_{\mathrm{obs}}\) averages \WSEITPConsistency{} for ITP, \WSEDraCorConsistency{} for DraCor, and \WSESmallvilleConsistency{} for Smallville. DraCor exceeds ITP by \WSEDraCorMinusITP{} points (95\% CI [\WSEDraCorMinusITPLow{}, \WSEDraCorMinusITPHigh{}]) and Smallville by \WSEDraCorMinusSmallville{} points ([\WSEDraCorMinusSmallvilleLow{}, \WSEDraCorMinusSmallvilleHigh{}]). Both \(C_{\mathrm{obs}}\) intervals include zero. Results under \(C_{\mathrm{adj}}\) are more favorable to DraCor and likewise do not indicate a DraCor disadvantage.

DraCor is not uniquely prone to internal contradiction over the evaluated horizon. Among inconsistent trajectories, the first contradiction appears at approximately step 10 in all three source families (Appendix Table~\ref{tab:supp-source-family-summary}). We also repeat the comparison after removing each of the seven DraCor plays in turn. The overall comparison remains stable regardless of which play is removed (Appendix Figure~\ref{fig:supp-source-family-leave-one-out}).

\subsection{Long-Horizon Error Patterns}
\label{sec:long-horizon-error-patterns}

  \paragraph{When contradictions appear}  A 10-step evaluation would miss 36.9\% of the contradictions eventually observed by step 20 in the Qwen group and 56.1\% in the Frontier group. Among cases with an agreed category, temporal conflicts are the most common in both groups. Identity and relationship conflicts tend to surface early against \texttt{InitialWorld} commitments, whereas character-state and world-fact conflicts more often arise later from the accumulated narrative history. The longer horizon thus exposes contradictions that short continuations systematically miss (Appendix~\ref{sec:horizon-sensitivity}; Table~\ref{tab:supp-inconsistency-reasons}).

 \paragraph{Semantic contradiction versus technical truncation}  Long-horizon breakdowns comprise semantic contradictions and technical truncations. A failure in a Character Agent, WMOD, or the Story Agent prevents the current narrative from being produced, whereas a WMEM failure follows a valid narrative and can truncate the subsequent trajectory.  A failed post-narration update can therefore preserve a coherent prefix while eliminating the future in which its consequences would have unfolded.  Some contradictions echo content introduced in Character-Agent proposals or WMOD plans, consistent with a conflict surviving multiple handoffs. Most cases, however, have no strong localized upstream match. The trace  identifies possible error pathways without assigning causal origins (Appendix~\ref{sec:error-propagation}).

 \section{Conclusion}

Evaluating stories as evolving trajectories reveals that sustained generation, canonical coherence, and meaningful development are distinct capacities. A trajectory can remain coherent by changing little, become richly eventful at the expense of accumulated canon, or end before either quality is tested over a long horizon. Keeping these outcomes separate shifts the question from whether a model can write a convincing continuation to whether it can sustain a world through consequential change.

More architecture is not necessarily more control. Greater scale chiefly extends how long generation lasts, while additional stages often shift trade-offs among the three outcomes rather than improving all of them. Components should earn their complexity by supplying information, coordination, or enforcement that downstream generation can actually use.

The source-family comparison does not support the simple expectation that divergence from a familiar plot necessarily destabilizes canon. More broadly, process evaluation must preserve both time and failure type: contradictions may emerge after substantial canon has accumulated, whereas component failures can end generation without producing contradictory text. The non-concave Frontier-model Pareto set reveals specialized Consistency- and Richness-favoring regimes rather than a smooth route to balance. The open challenge is to build systems whose stories can keep changing without losing what they have already become.

\small
\bibliographystyle{plainnat}
\bibliography{references}

\normalsize
\clearpage
\nobalance
\appendix
\twocolumn[
\begin{@twocolumnfalse}
\vspace*{-0.8em}
\begin{center}
{\fontsize{20}{23}\selectfont\bfseries\color{WSENavy}Appendix\par}
\end{center}
\vspace{0.75em}
\end{@twocolumnfalse}
]

\section{Benchmark Construction and Architecture}
\label{sec:episode-construction}

\subsection{Episode Construction and Cross-Architecture Comparison}

We standardized 17 \texttt{InitialWorld} records and selected the nine worlds in Table~\ref{tab:supp-initial-worlds}: two simulation-native settings and seven DraCor openings with self-contained five-character casts. Each world begins from one shared snapshot before player choices create a new timeline.

All nine worlds are converted to the same self-contained representation of characters, relationships, persistent facts, and time. Language and source family are not independently crossed, so family contrasts describe the worlds studied here rather than causal language effects.

Episode construction enumerates connected, self-contained five-character casts and rotates each character into the player role. A fixed selection procedure balances source family, focal-player representation, and relationship structure. This produces 42 \texttt{EpisodeSeed} records, with 14 per family. The native-language opening and three initial actions are then constructed without using any evaluated model.

At each step, a fixed sequence sampled uniformly over the three option positions determines the selected choice. All 12 conditions share this sequence and identical first-step option text, although the meaning of later options can diverge with the trajectory. Architecture may therefore affect the narrative, later options, selected actions, and subsequent continuation. The comparison captures the total effect of an architecture as the story evolves. It is not a direct module effect holding the narrative state fixed.

\subsection{WMOD as Pre-Narration Adjudication}

At step \(t\), the stateless World Model receives the selected player attempt \(P_t\), any Character-Agent intents \(I_t\) produced under \(B_2\) or \(B_3\), the canonical history \(H_t\), and structured state \(M_t\) when WMEM is enabled. It returns a \texttt{PlotPlan} \(W_t\) rather than reader-facing prose. The plan contains at most 12 \texttt{PlotBeat} records; each specifies \texttt{beat\_id}, \texttt{depends\_on}, \texttt{source\_intent\_ids}, and \texttt{content}. At least one beat traces to the player attempt, while background intents may be merged, delayed, prevented, or omitted. With WMOD enabled, the Story Agent receives \(P_t\) and \(W_t\) in place of the raw Character-Agent intents. The plan remains operational state until the Story Agent realizes it in Narrative \(N_t\), the canon-committing output.

\subsection{WMEM with Full-History Access}

After each narrative step, structured world memory records current character conditions, relationships, persistent facts, and supporting characters. A rule-based updater applies explicit model-proposed changes without adding inferred facts.
Concretely, the World Memory Agent emits a \texttt{MemoryDelta} of persistent changes, and the rule-based reducer applies those operations atomically to produce \(M_{t+1}\). For the next step, each Character Agent receives a deterministic actor-centric \texttt{CharacterMemory} projection of the global structured state.
Unchanged entries persist, and the memory cannot rewrite the generated narrative. Evaluators reconstruct canon independently from the narrative trajectory, so structured memory is never ground truth.

Let \(\mathcal{H}\) denote the policy of exposing the complete narrative history through the current step, and let \(M\) denote the structured current-state projection path. All evaluated conditions keep \(\mathcal{H}\) active and vary only the additional memory path. The relevant contrast is therefore
\begin{equation}
    \Delta_{M\mid \mathcal{H}}=Y(\mathcal{H},M)-Y(\mathcal{H},\varnothing),
\end{equation}
averaged over the stated bases and WMOD levels. This contrast measures the contribution of structured current-state information alongside the complete narrative history. The implemented WMEM path consists of extraction and rule-based reduction; retrieval, provenance verification, and conflict enforcement are outside that path. Its effect includes changes to later model inputs and the possible coverage loss caused by another model call.

\begin{table*}[t]
\centering
\footnotesize
\setlength{\tabcolsep}{4pt}
\renewcommand{\arraystretch}{0.96}
\begin{tabular}{@{}lllp{0.49\textwidth}r@{}}
\toprule
Source & Initial world & Language & Opening snapshot & Seeds \\
\midrule
ITP & Bianjing, 1100 CE & Chinese & Bianjing at the opening state in the year 1100 CE. & 14 \\
Smallville & 25-agent Smallville & English & Base simulation at midnight on February 13, 2023. & 14 \\
\addlinespace
DraCor & \textit{A Doll's House} & Dano--Norwegian & Act I, after the nurse brings the children upstairs. & 2 \\
DraCor & \textit{Uncle Vanya} & Russian & Act I, just after Serebryakov, Yelena, and Sonya return from their walk. & 2 \\
DraCor & \textit{Three Sisters} & Russian & Act I, after Andrei joins the guests at Irina's name-day gathering. & 2 \\
DraCor & \textit{The Cherry Orchard} & Russian & Act I, immediately after the family's arrival in the nursery. & 2 \\
DraCor & \textit{Hamlet} & English & Act I, Scene 2, after the court assembles at Elsinore. & 2 \\
DraCor & \textit{King Lear} & English & Act I, Scene 1, before the division of the kingdom is resolved. & 2 \\
DraCor & \textit{The Father} & Swedish & Opening of Act I, with the Captain, Laura, and Bertha at home. & 2 \\
\bottomrule
\end{tabular}
\caption{Initial worlds used to construct the 42 \texttt{EpisodeSeed} records. The snapshot identifies the shared canonical starting point, not a plot that generated trajectories are expected to follow.}
\label{tab:supp-initial-worlds}
\end{table*}

\begin{figure*}[t]
\centering
\fbox{%
\begin{minipage}{0.94\textwidth}
\small
\begin{align*}
(H_t,M_t^{(1)}),\ldots,(H_t,M_t^{(4)})
&\longrightarrow
\{\mathrm{CA}^{(1)}_t,\ldots,\mathrm{CA}^{(4)}_t\}_{\parallel}
\longrightarrow I_t, \\
(P_t,I_t,H_t,M_t)
&\longrightarrow \mathrm{WM}_t
\longrightarrow W_t, \\
(P_t,W_t,H_t,M_t)
&\longrightarrow \mathrm{SA}_t
\longrightarrow N_t, \\
(M_t,N_t)
&\longrightarrow \mathrm{MA}_t
\longrightarrow \Delta M_t
\xrightarrow{\mathrm{reduce}} M_{t+1}, \\
(H_t,N_t)&\longrightarrow H_{t+1},
\qquad (H_{t+1},M_{t+1})\circlearrowright \text{step }t+1.
\end{align*}
\vspace{-1.5ex}
\end{minipage}}
\caption{Per-step information flow for \(B_3+\mathrm{WMOD}+\mathrm{WMEM}\). \(H_t\) is canonical history, \(M_t\) structured world memory, and \(M_t^{(k)}\) the actor-specific view. \(P_t\), \(I_t\), \(W_t\), \(N_t\), and \(\Delta M_t\) denote the player intent, Character-Agent intents, World-Model plan, Story-Agent narrative, and memory update. Character Agents do not observe \(P_t\), and only \(N_t\) becomes part of canon.}
\label{fig:supp-factorized-information-path}
\end{figure*}

The Story Agent is the sole narrator in all conditions. Character Agents and the World Model are stateless at each step. Only the final narrative establishes canon. Structured memory is updated from that narrative, while unexecuted plans and proposed actions establish no story facts. In the experimental configuration, \(B_2\) uses one Character Agent to propose actions for all four non-player characters, whereas \(B_3\) uses four isolated calls to propose one action per non-player character. WMOD may reconcile these proposals into a plan of at most 12 ordered beats, but only the final narrative enters canon.

\section{Evaluation and Validation}
\label{sec:evaluation-definitions}

\subsection{Evaluator Blinding}

Each evaluator receives an independently prepared view containing the \texttt{InitialWorld}, opening, language, and generated trajectory. The view excludes model and architecture identities, source-family labels, internal outputs, memory, failures, costs, and the other evaluator's result. DeepSeek-V4-Pro appears both among the evaluated models and in the Judge ensemble, but its identity is hidden during evaluation. Character names and canon remain visible as necessary evidence and may reveal recognizable works.

\subsection{Consistency Under Unequal Coverage}
\label{sec:two-consistency-estimands}

For trajectory \(i\), let \(L_i\) be its number of narrative steps available for evaluation and set \(T_{\mathrm{inc},i}=\infty\) when no contradiction occurs in that prefix. The exact step-\(k\) risk set and first-event count are
\begin{equation}
\begin{aligned}
Y_k&=\sum_i \mathbb{I}[L_i\geq k,\,T_{\mathrm{inc},i}\geq k],\\
D_k&=\sum_i \mathbb{I}[L_i\geq k,\,T_{\mathrm{inc},i}=k].
\end{aligned}
\end{equation}
A trajectory with its first contradiction at step \(k\) belongs to both \(Y_k\) and \(D_k\) at that step, then leaves later risk sets. A contradiction-free trajectory contributes through \(L_i\) and is then right-censored. Thus \(D_k\) counts first events at step \(k\), not all contradictions observed up to that step.

For the experiments reported here, \(H=20\). The estimated contradiction-free curve and the two Consistency summaries are
\begin{equation}
\begin{aligned}
\widehat S_C(t)
&=\prod_{k\leq t:\,Y_k>0}\left(1-\frac{D_k}{Y_k}\right),\\
C_{\mathrm{obs}}
&=\frac{100}{H}\sum_{t=1}^{H}\widehat S_C(t),\\
C_{\mathrm{adj}}
&=\frac{100}{H}\sum_{t=1}^{H}\mathbb{I}[Y_t>0]\widehat S_C(t).
\end{aligned}
\end{equation}

Both summaries use the same verdicts, first-contradiction events, and censoring. They differ only after no trajectory remains at risk. \(C_{\mathrm{obs}}\) carries forward the last estimable survival value, while \(C_{\mathrm{adj}}\) assigns zero after support is exhausted. The heatmap pairs \(C_{\mathrm{obs}}\) with a separate generation-coverage axis. The Pareto view uses \(C_{\mathrm{adj}}\). Neither summary recovers unseen continuations or treats generation failure as a contradiction.

\begin{figure*}[t]
\centering
\includegraphics[width=\textwidth]{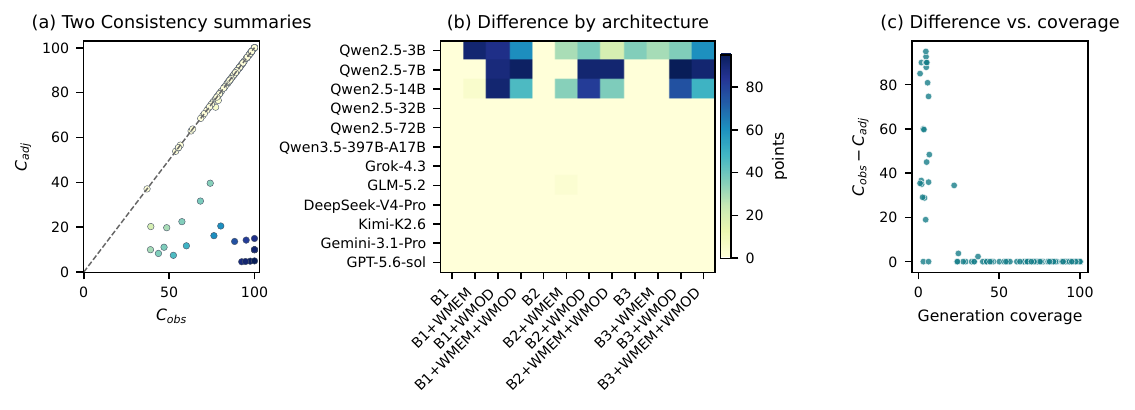}
\caption{Comparison of the two Consistency summaries across all \WSECellCount{} model--architecture cells. Panel (a) compares \(C_{\mathrm{obs}}\) with \(C_{\mathrm{adj}}\). Panel (b) shows where they differ across architectures, and panel (c) relates that difference to Generation Coverage \(G\). The difference is concentrated in lower-coverage Qwen cells and changes no observed judgment or event location.}
\label{fig:supp-consistency-estimands}
\end{figure*}

Figure~\ref{fig:supp-consistency-estimands} shows that the two values coincide in 119 of 144 cells. Their differences occur mainly in Qwen conditions with few trajectories remaining at later steps and are nearly absent among Frontier models. Recomputing non-dominance with either definition leaves the same six-cell Pareto set across all 144 cells. The Qwen-only sets differ by one member, so we continue to report both summaries for Qwen comparisons affected by sparse later-step evidence.

\subsection{Supportedness of the Empirical Pareto Set}
\label{sec:pareto-supportedness}

Let \(\mathcal{Y}_F\) be the 72 Frontier model--architecture point estimates. A non-dominated point \(y_i\) is \emph{supported} if some strictly positive weight vector selects it under a linear scalarization,
\begin{equation}
    y_i\in\arg\max_{y\in\mathcal{Y}_{F}}\mathbf{w}^{\top}y,
    \qquad \mathbf{w}\in\mathbb{R}^{2}_{>0}.
\end{equation}
Only two endpoints are supported: DeepSeek-V4-Pro B3+WMOD and Gemini-3.1-Pro B3+WMEM+WMOD. In increasing Consistency order, the four intermediate non-dominated cells fall 2.84, 8.26, 8.15, and 0.05 Richness points below the chord joining these endpoints. The final gap is negligible relative to sampling uncertainty, so this classification describes the estimated frontier only.

\subsection{Richness Annotation}
\label{sec:richness-computation}

Let \(d\) be the number of non-repetitive development units per observed narrative step, and let \(d_{95}=3.5\) be the 95th percentile in the calibration data. For each breadth dimension \(j\), \(s_j\in\{1,\ldots,5\}\) is its anchored rating. The exact computation is
\begin{equation}
\begin{aligned}
Q
&=\operatorname{clip}\!\left(\frac{\log(1+d)}{\log(1+d_{95})},0,1\right),\\
B
&=\frac{1}{3}\sum_{j\in\{\mathrm{event},\mathrm{character},\mathrm{world}\}}
\frac{s_j-1}{4},\\
R&=100\sqrt{QB}.
\end{aligned}
\end{equation}
Here \(\operatorname{clip}(x,0,1)\) truncates \(x\) to \([0,1]\), and the character and world labels refer to character and relationship development and world or social expansion.

\subsection{Human Validation of Judges}
\label{sec:evaluation-validation}

Consistency and Richness use separate blinded human-reference samples. Annotators receive the trajectory and rubric, but not model, architecture, sampling stratum, or model-Judge outputs.

\begin{figure*}[!t]
\centering
\includegraphics[width=0.77\textwidth]{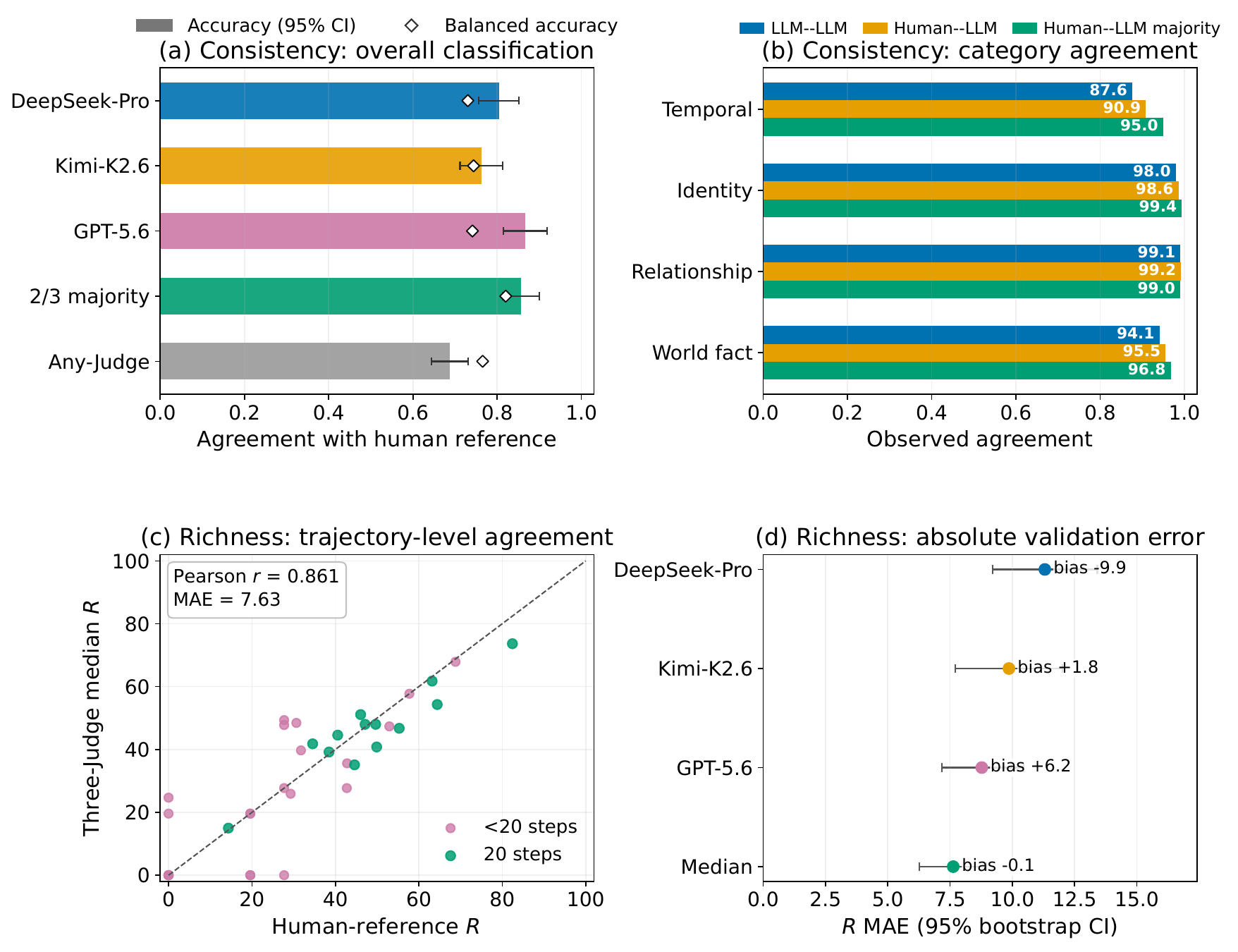}
\caption{Human validation of the Judges. (a) Design-weighted Consistency classification on 120 trajectories. (b) One-vs-rest observed agreement across the same sample; the pairwise series average three comparisons. (c--d) Richness agreement and error on a separate 36-trajectory sample.}
\label{fig:supp-human-validation}
\end{figure*}

Figure~\ref{fig:supp-human-validation} summarizes both validation studies. The Consistency sample draws 120 human-labeled cases from 2,447 trajectories with no uncertain Judge verdicts. Inverse-probability weights restore the target distribution, while 14 uncertain cases are kept only for additional analysis. Panel (b) treats each represented inconsistency type as a one-vs-rest judgment over all 120 trajectories, so shared presence and shared absence both count as agreement.

The separate Richness sample contains 36 cases stratified by six models and three source families. The three-Judge median provides the closest aggregate match to the human reference, with the lowest MAE and near-zero bias. Across the three breadth dimensions used in \(R\), its ordinal ratings fall within one point of the human reference in 97.8\% of weighted comparisons (mean quadratic-weighted \(\kappa=.780\)). Together, the two studies support the aggregate Judge measures used in the main analysis.

\subsection{Long-Horizon Sensitivity}
\label{sec:unequal-coverage}
\label{sec:horizon-sensitivity}

A shorter trajectory has fewer opportunities to contradict itself or develop. First-contradiction analysis remains at risk until the event or the end of the evaluated prefix, and the trajectories in these experiments end at step 20. A missing continuation is not counted as a contradiction. Richness is instead weighted by observed narrative steps because, unlike Consistency, it is not determined by a single first event.

For each shorter horizon, we recompute Generation Coverage \(G_h\) and \(C_{\mathrm{obs}}(h)\) from the same trajectories and give the six models equal weight within each group.

\begin{figure*}[t]
\centering
\includegraphics[width=0.72\textwidth]{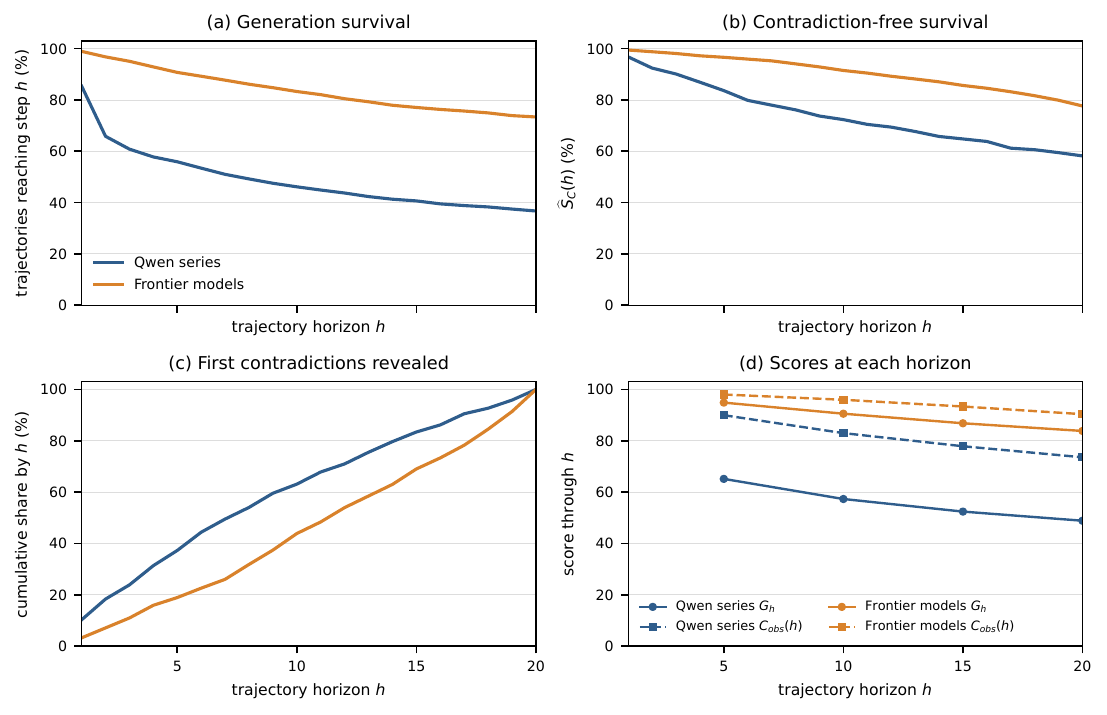}
\caption{Horizon sensitivity for the two six-model groups. Panels (a--b) show equal-model mean generation and contradiction-free survival. Panel (c) reports the timing distribution of first contradictions observed by step 20. Panel (d) recomputes the normalized summaries at \(h=5,10,15,20\).}
\label{fig:supp-horizon-sensitivity}
\end{figure*}

Figure~\ref{fig:supp-horizon-sensitivity} shows that only 63.1\% of Qwen-series and 43.9\% of Frontier first contradictions observed by step 20 had appeared by step 10. A ten-step evaluation would therefore miss 36.9\% and 56.1\%, respectively, of the first contradictions eventually observed in these same trajectories.

Because Richness Judges assess each observed trajectory as a whole, we do not reuse those ratings to estimate Richness at earlier horizons.

\section{Architecture Contrasts and Robustness}
\label{sec:generation-coverage}
\label{sec:architecture-contrasts}

\label{sec:error-propagation}
Architecture affects both story quality and the opportunity to reach later steps. Figure~\ref{fig:supp-compute-contrasts} therefore shows computational cost alongside Generation Coverage. Extra computation alone does not provide more narrative evidence. Figure~\ref{fig:supp-generation-failures} also separates valid story endings from generation failures instead of treating the latter as Consistency or Richness outcomes.

The directed trace in Figure~\ref{fig:supp-factorized-information-path} distinguishes generation failures from possible narrative precursors. Character-Agent, World-Model, and Story-Agent failures prevent the current narrative. A World-Memory failure occurs after a valid narrative, so that step remains canon even though later generation stops. To trace possible precursors, we localize the first conflict and compare it with the player intent, Character-Agent intents, World-Model beats, and memory views. Conservative lexical matches identify possible precursors rather than causal origins.

Table~\ref{tab:supp-error-propagation-summary} separates generation failures from possible narrative precursors. Across both model groups, 1,709 generation failures occur before the current narrative is produced. The 1,021 World-Memory failures instead follow a valid narrative and prevent 12,672 later narrative steps from being produced. Added structure can therefore reduce sustained generation through extra handoffs without contradicting text already shown.

Temporal conflicts are the most common category on which Judges agree in both model groups. Identity and relationship conflicts tend to appear earlier and usually contradict \texttt{InitialWorld} commitments. Character-state and world-fact conflicts more often contradict earlier narratives. Strong Character-Agent precursors appear in 89 inconsistent trajectories, including 42 that persist through the World Model. Another 27 trajectories have only a World-Model precursor. Most cases lack a strong localized upstream match. The trace thus shows that errors can pass through multiple layers, but not that added layers are their predominant source.

\begin{table}[!t]
\centering
\scriptsize
\setlength{\tabcolsep}{2pt}
\renewcommand{\arraystretch}{1.04}
\begin{tabular*}{\columnwidth}{@{\extracolsep{\fill}}p{0.62\columnwidth}rr@{}}
\toprule
Quantity & Qwen & Frontier \\
\midrule
\multicolumn{3}{@{}l}{\textit{Generation failures}} \\
Centralized Character Agent failures & 1.1\% & 5.4\% \\
Distributed Character Agent failures & 1.9\% & 12.8\% \\
World Model failures & 33.6\% & 6.9\% \\
Story Agent failures & 24.4\% & 4.8\% \\
World Memory Agent failures & 42.7\% & 24.8\% \\
\midrule
\multicolumn{3}{@{}l}{\textit{Possible narrative precursors}} \\
Memory precursor & 0.8\% & 0.0\% \\
Player-intent precursor & 3.2\% & 0.4\% \\
Character-Agent precursor & 12.2\% & 5.8\% \\
World-Model precursor & 2.8\% & 2.8\% \\
No strong recorded precursor & 55.8\% & 64.1\% \\
Violation not localized & 9.7\% & 12.7\% \\
Plurality tie & 15.6\% & 14.2\% \\
\bottomrule
\end{tabular*}
\caption{Two complementary error-propagation audits. Generation-failure rates use trajectories in which the component is enabled. Precursor rates use majority-inconsistent trajectories and report the plurality origin of conservative lexical matches to upstream content. These matches identify possible precursors, not causal effects or errors repaired downstream.}
\label{tab:supp-error-propagation-summary}
\end{table}

\subsection{Per-Model Architecture Effects}

Table~\ref{tab:supp-model-effects-all-models}, the per-model counterpart to main-text Table 1, uses the same paired measures and 10,000 hierarchical source-cluster/\texttt{EpisodeSeed} bootstrap draws. Its intervals describe seed variation among the models studied. They do not capture decoding or model-population uncertainty and are not adjusted for multiple comparisons.

\subsection{Base-Conditional Module Effects}

Table~\ref{tab:supp-model-conditional-all-models} compares each module on versus off within a base while averaging over the other module. These conditional contrasts compare the resulting trajectories as a whole rather than holding the narrative state fixed.

The decomposition shows no base with a stable WMEM Consistency gain. WMOD has the clearest topology dependence: Frontier Richness changes by \(+0.4\;[-1.3,+2.1]\) at B1, \(+4.3\;[+2.6,+6.0]\) at B2, and \(+5.3\;[+4.0,+6.6]\) at B3.

\subsection{Full-Horizon Paired Comparison}
\label{sec:full-horizon-paired}

This sensitivity retains an architecture pair only when both conditions reach step 20 with valid scores. It then averages paired differences within each model and the available model estimates within each group.

\begin{figure*}[t]
\centering
\includegraphics[width=\textwidth]{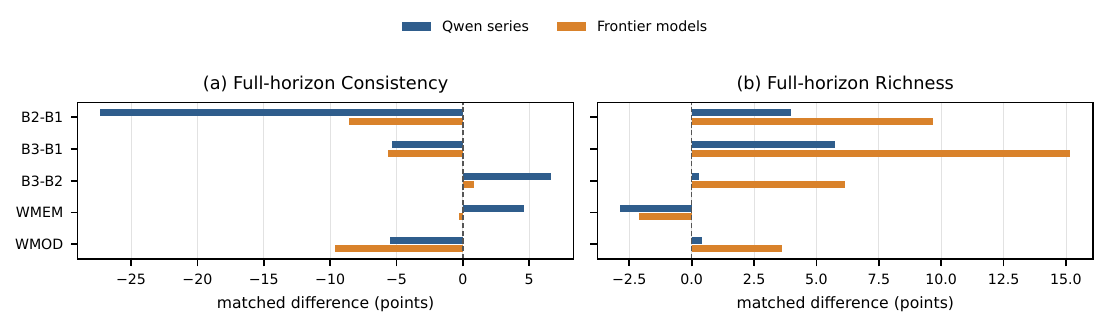}
\caption{Architecture contrasts among paired conditions in which both trajectories reach the full horizon. Bars average the six model-specific estimates in each group. Positive values favor the first-named base or module-on condition. This analysis complements the primary analysis, which includes each trajectory up to its observed length.}
\label{fig:supp-full-horizon-paired}
\end{figure*}

The paired results in Figure~\ref{fig:supp-full-horizon-paired} preserve the character-agent trade-off and the absence of a joint WMEM gain, showing that the main pattern is not created solely by comparing short and full-horizon prefixes.

\begingroup
\begin{figure*}[t]
\centering
\includegraphics[width=\textwidth]{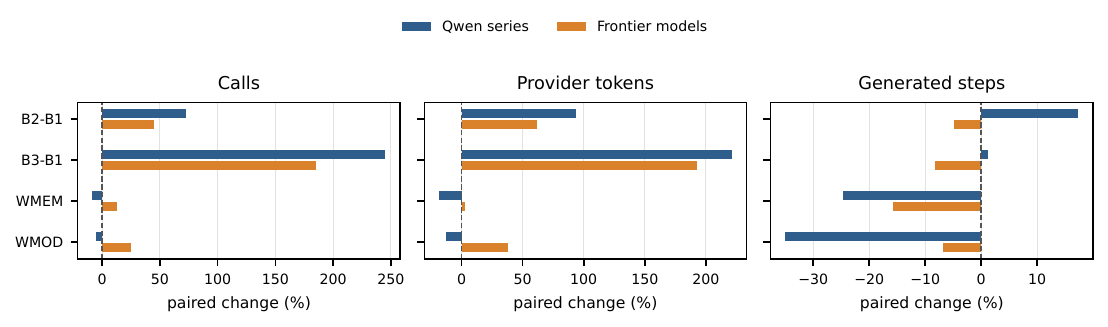}
\caption{Generation cost and coverage under four architecture contrasts. Bars average within-model percentage changes across the six models in each group. Positive values favor the first-named base or module-on condition. B3 invokes four isolated Character Agents per active step, and lower coverage can reduce later calls and tokens.}
\label{fig:supp-compute-contrasts}
\end{figure*}
\endgroup

\section{Model and Scale Effects}
\label{sec:model-scale-analysis}

The models exhibit distinct profiles across Generation Coverage, Consistency, and Richness, with no common ranking across the three outcomes. This model-level variation is separate from scaling, which we assess only among the comparable dense Qwen2.5 checkpoints. Figure~\ref{fig:supp-generation-failures} shows how often each model sustains generation through the 20-step horizon. Figure~\ref{fig:supp-qwen-survival} then shows the corresponding Qwen trajectories step by step.

\begingroup
\begin{figure*}[t]
\centering
\includegraphics[width=0.97\textwidth]{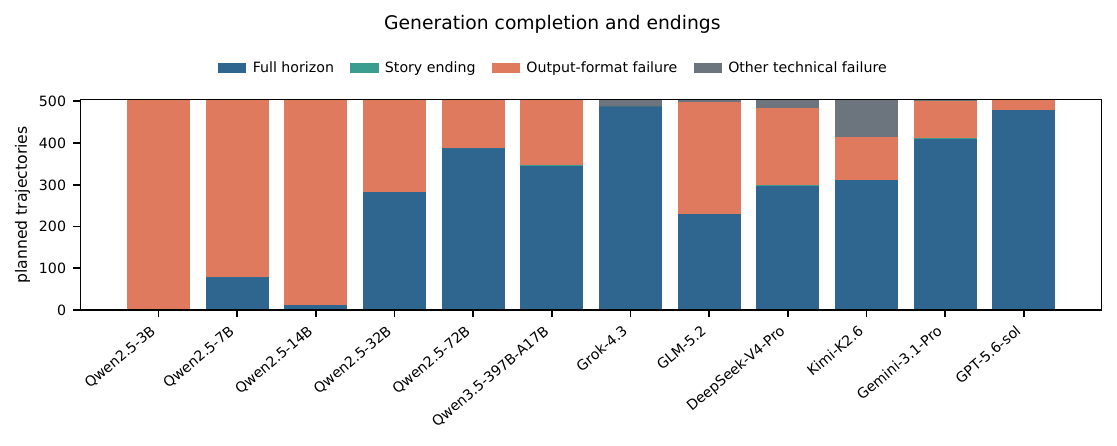}
\caption{Generation outcomes by model across all 12 architectures. Each bar represents 504 planned trajectories. \emph{Full horizon} indicates 20 completed narrative steps. \emph{Story ending} denotes an accepted ending, such as the player's death, permanent incapacity, or a genuinely closed story. Only four occur. Output-format and other technical failures reduce Generation Coverage but are not story outcomes.}
\label{fig:supp-generation-failures}
\end{figure*}
\endgroup

\begingroup
\begin{figure*}[!t]
\centering
\includegraphics[width=\textwidth]{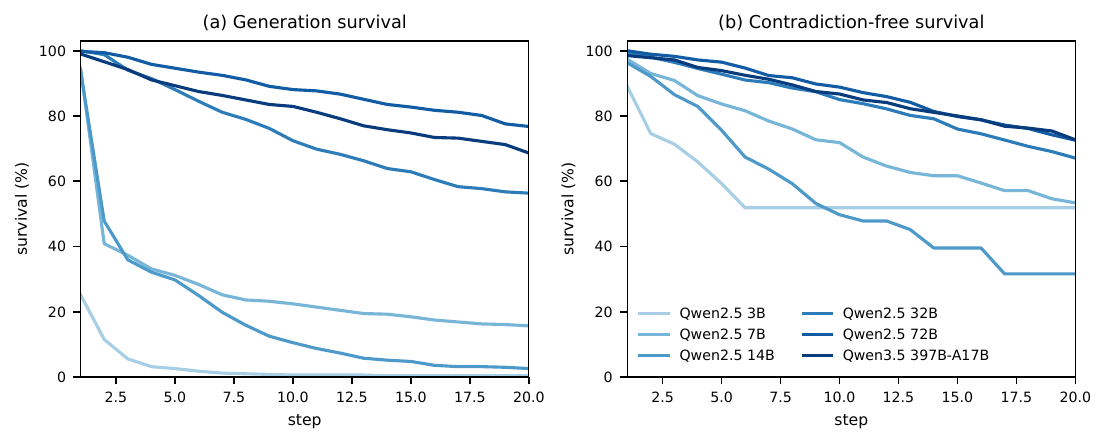}
\caption{Finite-horizon survival curves for the Qwen series. Panel (a) shows the probability of generating through step \(t\). Panel (b) shows contradiction-free survival, with trajectories that stop for technical reasons right-censored. Dotted tails have fewer than 30 trajectories at risk. Curves end at the experimental horizon \(H=20\), and their integrals yield the summaries in Figure~\ref{fig:supp-qwen-summary}. Qwen3.5-397B-A17B is shown separately as an MoE model.}
\label{fig:supp-qwen-survival}
\end{figure*}
\endgroup

\subsection{Qwen-Series Scaling}

Figure~\ref{fig:supp-qwen-summary} condenses the step-resolved curves into finite-horizon model summaries. Dense fits use \(\log_2\) total parameters for the five Qwen2.5 checkpoints. Qwen3.5-397B-A17B is excluded because total parameter count cannot be compared directly between its sparse MoE architecture and the dense models.

\begingroup
\begin{figure}[!t]
\centering
\includegraphics[width=\columnwidth]{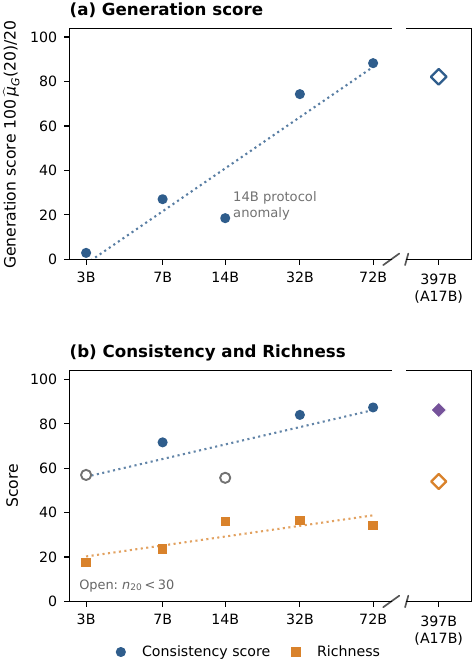}
\caption{Qwen-series scaling under unequal trajectory coverage. Dense-model fits use log parameter count for Qwen2.5 3B--72B. Qwen3.5-397B-A17B is shown separately as an MoE model. Open markers have fewer than 30 trajectories at risk at step 20.}
\label{fig:supp-qwen-summary}
\end{figure}
\endgroup

\begingroup
\noindent\begin{minipage}{\columnwidth}
\centering
\small
\begin{tabular}{@{}lrrr@{}}
\toprule
Outcome & \(\alpha\) & \(\beta\) & \(R^2\) \\
\midrule
Generation Coverage \(G\) & -32.62 & 19.32 & .879 \\
\(G\), excluding 14B & -26.46 & 19.17 & .992 \\
Consistency \(C_{\mathrm{obs}}\) & 45.71 & 6.55 & .639 \\
Richness \(R\) & 13.86 & 4.03 & .720 \\
\bottomrule
\end{tabular}
\captionsetup{hypcap=false}
\captionof{table}{Model-level fits \(Y=\alpha+\beta\log_2P\), where \(P\) is the dense-model parameter count in billions. The primary rows use five Qwen2.5 models. The sensitivity row omits 14B, and Qwen3.5 MoE is excluded. The fitted associations apply to the five studied dense Qwen2.5 checkpoints.}
\label{tab:supp-descriptive-scaling}
\end{minipage}
\endgroup

Table~\ref{tab:supp-descriptive-scaling} shows that the unusually low Generation Coverage at 14B does not drive its slope: excluding that checkpoint changes the estimate from 19.32 to 19.17 points per parameter doubling. Consistency and Richness vary non-monotonically across the five model sizes.

Tables~\ref{tab:supp-inconsistency-reasons} and~\ref{tab:supp-inconsistency-prior-source} expand the long-horizon error analysis by separating the first-conflict categories from the location of the prior canonical commitment.

\begingroup
\begin{table}[!t]
\centering
\scriptsize
\setlength{\tabcolsep}{2.3pt}
\begin{tabular*}{\columnwidth}{@{\extracolsep{\fill}}lcc@{}}
\toprule
First-violation reason & Qwen: $n$ (\%) / step & Frontier: $n$ (\%) / step \\
\midrule
Temporal conflict & 131 (25.8\%) / 8.0 & 144 (31.0\%) / 11.0 \\
World-fact conflict & 52 (10.3\%) / 10.0 & 88 (18.9\%) / 13.0 \\
Character-state conflict & 70 (13.8\%) / 7.0 & 38 (8.2\%) / 9.0 \\
Identity conflict & 42 (8.3\%) / 2.0 & 23 (4.9\%) / 7.0 \\
Relationship conflict & 34 (6.7\%) / 2.0 & 6 (1.3\%) / 4.0 \\
Knowledge conflict & 2 (0.4\%) / 10.0 & 4 (0.9\%) / 11.5 \\
Causal conflict & 0 (0.0\%) / -- & 0 (0.0\%) / -- \\
Other & 2 (0.4\%) / 2.0 & 1 (0.2\%) / 15.0 \\
No 2-Judge agreement & 174 (34.3\%) / 10.0 & 160 (34.4\%) / 13.0 \\
\bottomrule
\end{tabular*}
\caption{Adjudicated reasons for the first inconsistency, excluding generation failures. A category requires agreement from at least two inconsistent Judges. Percentages use all majority-inconsistent trajectories in each model panel. The value after each slash is the median first contradictory step.}
\label{tab:supp-inconsistency-reasons}
\end{table}

\endgroup

\begingroup
\begin{table}[!t]
\centering
\scriptsize
\setlength{\tabcolsep}{1.2pt}
\renewcommand{\arraystretch}{1.10}
\begin{tabular*}{\columnwidth}{@{\extracolsep{\fill}}lrrrrrr@{}}
\toprule
\shortstack[c]{Agreed\\reason} & \shortstack[c]{Initial\\World} & \shortstack[c]{Episode\\opening} & \shortstack[c]{Earlier\\Narrative} & \shortstack[c]{Current\\Narrative} & \shortstack[c]{Unresolved\\tied} & \raisebox{4pt}{Total} \\
\midrule
Temporal & 21 & 0 & 82 & 17 & 155 & 275 \\
World fact & 2 & 0 & 63 & 2 & 73 & 140 \\
Character state & 5 & 0 & 54 & 9 & 40 & 108 \\
Identity & 48 & 0 & 0 & 1 & 16 & 65 \\
Relationship & 39 & 0 & 0 & 0 & 1 & 40 \\
Knowledge & 0 & 0 & 3 & 1 & 2 & 6 \\
\bottomrule
\end{tabular*}
\caption{Candidate source of the canonical commitment violated at the first inconsistent step, pooled across both model panels. Sources are localized from the Judges' stated prior commitment. Ambiguous and tied cases are grouped as unresolved. The table identifies where a commitment was recorded, not which agent caused the conflict.}
\label{tab:supp-inconsistency-prior-source}
\end{table}

\endgroup

\section{Source-Family Inference}
\label{sec:source-family-analysis}

Table~\ref{tab:supp-source-family-summary} uses 10,000 hierarchical bootstrap draws over source clusters and \texttt{EpisodeSeed} records while retaining models and paired architectures. Both primary \(C_{\mathrm{obs}}\) DraCor contrasts include zero, whereas the \(C_{\mathrm{adj}}\) sensitivity is more favorable to DraCor. In the leave-one-play-out analysis, all seven DraCor--Smallville contrasts remain positive. Six of the seven DraCor--ITP contrasts remain positive; omitting one play gives a difference of \(-0.27\) points (Figure~\ref{fig:supp-source-family-leave-one-out}).

\begingroup
\begin{figure}[!t]
\centering
\includegraphics[width=\columnwidth]{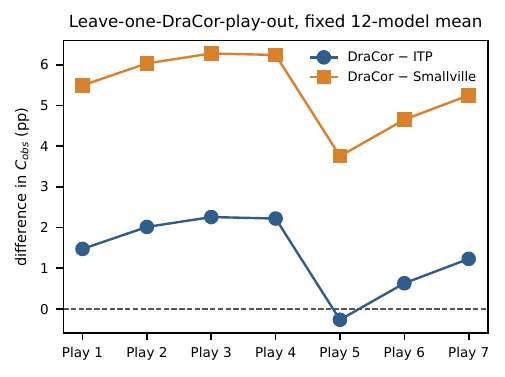}
\caption{Leave-one-DraCor-play-out sensitivity for the 12-model \(C_{\mathrm{obs}}\) contrasts. Each point compares DraCor with a simulation-native control after omitting one play. All DraCor--Smallville contrasts remain positive; one DraCor--ITP contrast is slightly below zero (\(-0.27\) points).}
\label{fig:supp-source-family-leave-one-out}
\end{figure}
\endgroup

\begin{table}[!t]
\centering
\begingroup
\scriptsize
\setlength{\tabcolsep}{2.4pt}
\renewcommand{\arraystretch}{1.04}
\begin{tabular*}{\columnwidth}{@{\extracolsep{\fill}}lccc@{}}
\toprule
Measure & ITP & DraCor & Smallville \\
\midrule
\multicolumn{4}{@{}l}{\textit{Consistency: estimate [95\% bootstrap CI]}} \\
$C_{\mathrm{obs}}$ & 82.1 [78.4, 86.1] & 83.5 [79.1, 87.9] & 78.1 [75.7, 84.0] \\
$C_{\mathrm{adj}}$ & 80.3 [77.3, 82.8] & 83.5 [77.1, 87.8] & 75.2 [72.4, 77.0] \\
\addlinespace
\multicolumn{4}{@{}l}{DraCor \textit{contrast: difference [95\% bootstrap CI]}} \\
$C_{\mathrm{obs}}$ & +1.4 [-4.6, +6.9] & --- & +5.4 [-2.2, +9.9] \\
$C_{\mathrm{adj}}$ & +3.2 [-3.8, +8.3] & --- & +8.3 [+1.8, +13.5] \\
\addlinespace
\multicolumn{4}{@{}l}{\textit{Realized temporal exposure}} \\
Observed narrative steps & 12.5 & 13.3 & 14.0 \\
Day advance (\%) & 0.7 & 0.3 & 1.0 \\
Full horizon (\%) & 50.4 & 54.0 & 60.8 \\
First contradiction step & 9.8 & 10.2 & 9.6 \\
Relative day at contradiction & 1.1 & 1.1 & 1.2 \\
\bottomrule
\end{tabular*}
\endgroup

\caption{Source-family Consistency and trajectory timing. Cells report equal-model means with 95\% hierarchical bootstrap intervals. DraCor contrasts subtract ITP or Smallville from DraCor.}
\label{tab:supp-source-family-summary}
\end{table}

Consistency is judged against the supplied \texttt{InitialWorld} and emerging trajectory, not against a published continuation. Prior knowledge is observable only when it creates an internal contradiction.

As Table~\ref{tab:supp-source-family-summary} shows, all families use the same relative-day representation and record whether the story advances to a new day. Such advances are rare and least common in DraCor, while mean first-contradiction steps are similar. The similar timing gives no indication that the relative-day representation drives the family pattern.

\clearpage
\nobalance

\begin{table*}[p]
\centering
\scriptsize
\setlength{\tabcolsep}{2.4pt}
\renewcommand{\arraystretch}{0.88}
\begin{tabular}{@{}llrrrr@{}}
\toprule
Generator & Contrast & $\Delta G$ & $\Delta C_{\mathrm{obs}}$ & $\Delta C_{\mathrm{adj}}$ & $\Delta R$ \\
\midrule
\multicolumn{6}{@{}l}{\textbf{Qwen series}} \\
\textbf{Qwen2.5-3B} & B2 -- B1 & $+1.8\,[-0.8, +4.4]$ & $-37.6\,[-64.2, -8.4]$ & $+0.1\,[-24.2, +33.3]$ & $+5.8\,[-1.1, +15.0]$ \\
 & B3 -- B1 & $+0.0\,[-1.7, +1.3]$ & $-38.9\,[-62.0, -9.8]$ & $-20.0\,[-24.3, +6.8]$ & $+10.8\,[-2.3, +20.2]$ \\
 & B3 -- B2 & $-1.8\,[-3.7, -0.3]$ & $-1.3\,[-29.1, +31.6]$ & $-20.0\,[-30.6, +3.2]$ & $+5.0\,[-10.0, +13.9]$ \\
 & WMEM on -- off & $+0.7\,[-1.1, +2.3]$ & $-9.3\,[-33.8, +10.6]$ & $-25.1\,[-33.5, +5.6]$ & $+1.2\,[-5.3, +8.0]$ \\
 & WMOD on -- off & $-0.8\,[-2.5, +0.7]$ & $-1.2\,[-25.0, +27.1]$ & $-20.0\,[-29.9, +9.0]$ & $-4.9\,[-10.4, +3.5]$ \\
\addlinespace[2pt]
\textbf{Qwen2.5-7B} & B2 -- B1 & $+4.6\,[+0.2, +9.0]$ & $+7.6\,[-2.8, +18.3]$ & $+7.7\,[-3.4, +17.5]$ & $-2.2\,[-4.7, +0.3]$ \\
 & B3 -- B1 & $+1.5\,[-2.8, +5.7]$ & $+8.3\,[-1.9, +19.3]$ & $+7.1\,[-3.2, +17.6]$ & $+0.2\,[-2.1, +2.5]$ \\
 & B3 -- B2 & $-3.2\,[-7.4, +1.2]$ & $+0.7\,[-8.1, +9.8]$ & $-0.6\,[-8.8, +10.1]$ & $+2.4\,[+0.1, +4.6]$ \\
 & WMEM on -- off & $-22.4\,[-26.1, -18.5]$ & $+0.6\,[-6.1, +7.6]$ & $+0.7\,[-8.2, +6.9]$ & $-0.4\,[-2.6, +1.5]$ \\
 & WMOD on -- off & $-44.7\,[-49.2, -40.0]$ & $+26.0\,[+19.9, +31.9]$ & $-64.9\,[-71.3, -56.9]$ & $-18.8\,[-20.8, -16.8]$ \\
\addlinespace[2pt]
\textbf{Qwen2.5-14B} & B2 -- B1 & $+1.5\,[-1.5, +4.3]$ & $-3.0\,[-27.1, +21.6]$ & $-6.1\,[-12.9, +4.7]$ & $+6.7\,[+2.7, +10.6]$ \\
 & B3 -- B1 & $+1.9\,[-1.9, +5.7]$ & $-10.4\,[-33.0, +13.5]$ & $-6.4\,[-13.0, +6.3]$ & $+9.2\,[+5.7, +12.5]$ \\
 & B3 -- B2 & $+0.4\,[-2.0, +2.9]$ & $-7.4\,[-30.0, +14.2]$ & $-0.3\,[-7.8, +8.6]$ & $+2.5\,[-1.5, +6.7]$ \\
 & WMEM on -- off & $-7.3\,[-9.9, -4.7]$ & $-16.9\,[-31.4, +4.5]$ & $-3.9\,[-12.4, +1.7]$ & $-1.8\,[-5.0, +1.2]$ \\
 & WMOD on -- off & $-25.2\,[-30.4, -19.5]$ & $+14.8\,[-1.6, +34.9]$ & $-41.3\,[-48.4, -31.1]$ & $-12.0\,[-15.1, -9.1]$ \\
\addlinespace[2pt]
\textbf{Qwen2.5-32B} & B2 -- B1 & $+5.8\,[-0.7, +12.1]$ & $-8.6\,[-14.6, -2.8]$ & $-8.6\,[-14.9, -2.8]$ & $+3.2\,[+0.8, +5.8]$ \\
 & B3 -- B1 & $+4.4\,[-1.8, +10.6]$ & $-13.1\,[-19.0, -6.9]$ & $-13.1\,[-19.8, -6.9]$ & $+6.4\,[+3.5, +9.4]$ \\
 & B3 -- B2 & $-1.4\,[-5.9, +3.2]$ & $-4.5\,[-13.3, +4.4]$ & $-4.5\,[-13.9, +4.5]$ & $+3.2\,[+0.7, +5.8]$ \\
 & WMEM on -- off & $-38.4\,[-44.0, -32.8]$ & $-1.9\,[-7.9, +4.1]$ & $-1.9\,[-8.2, +3.8]$ & $-1.2\,[-3.2, +0.7]$ \\
 & WMOD on -- off & $-7.0\,[-12.1, -2.0]$ & $-1.4\,[-8.0, +5.2]$ & $-1.4\,[-7.7, +5.4]$ & $-2.4\,[-4.5, -0.4]$ \\
\addlinespace[2pt]
\textbf{Qwen2.5-72B} & B2 -- B1 & $+0.6\,[-3.6, +5.2]$ & $-3.4\,[-9.0, +1.9]$ & $-3.4\,[-9.0, +1.9]$ & $-0.8\,[-2.9, +1.3]$ \\
 & B3 -- B1 & $-0.2\,[-4.8, +4.5]$ & $-6.7\,[-12.5, -0.6]$ & $-6.7\,[-12.5, -0.6]$ & $-0.6\,[-3.3, +2.1]$ \\
 & B3 -- B2 & $-0.8\,[-4.7, +3.0]$ & $-3.3\,[-9.5, +3.2]$ & $-3.3\,[-9.5, +3.2]$ & $+0.2\,[-1.7, +2.0]$ \\
 & WMEM on -- off & $-18.5\,[-23.7, -13.5]$ & $-1.2\,[-5.3, +3.5]$ & $-1.2\,[-5.3, +3.5]$ & $-1.1\,[-3.2, +0.9]$ \\
 & WMOD on -- off & $-1.5\,[-4.9, +2.0]$ & $-4.7\,[-9.8, +0.4]$ & $-4.7\,[-9.8, +0.4]$ & $-3.8\,[-5.5, -1.8]$ \\
\addlinespace[2pt]
\textbf{Qwen3.5-397B-A17B} & B2 -- B1 & $-9.7\,[-14.9, -4.9]$ & $-10.2\,[-16.9, -3.9]$ & $-10.2\,[-16.9, -3.9]$ & $+11.6\,[+9.3, +13.8]$ \\
 & B3 -- B1 & $-14.0\,[-18.5, -9.7]$ & $-11.2\,[-19.1, -3.5]$ & $-11.2\,[-19.1, -3.5]$ & $+16.1\,[+13.4, +18.6]$ \\
 & B3 -- B2 & $-4.3\,[-9.6, +1.1]$ & $-1.0\,[-7.9, +5.5]$ & $-1.0\,[-7.9, +5.5]$ & $+4.5\,[+2.1, +6.9]$ \\
 & WMEM on -- off & $-22.0\,[-26.6, -17.5]$ & $-4.9\,[-9.9, +0.1]$ & $-4.9\,[-9.9, +0.1]$ & $-4.4\,[-6.0, -2.9]$ \\
 & WMOD on -- off & $-3.1\,[-7.0, +0.7]$ & $-5.4\,[-11.9, +0.7]$ & $-5.4\,[-11.9, +0.7]$ & $+6.2\,[+4.6, +7.8]$ \\
\addlinespace[2pt]
\addlinespace[2pt]
\multicolumn{6}{@{}l}{\textbf{Frontier models}} \\
\textbf{Grok-4.3} & B2 -- B1 & $+0.7\,[-1.4, +3.6]$ & $-4.9\,[-9.2, -1.1]$ & $-4.9\,[-9.2, -1.1]$ & $+6.1\,[+3.3, +9.0]$ \\
 & B3 -- B1 & $-0.4\,[-4.8, +4.5]$ & $-5.8\,[-10.9, -1.2]$ & $-5.8\,[-10.9, -1.2]$ & $+16.2\,[+13.1, +19.5]$ \\
 & B3 -- B2 & $-1.1\,[-4.1, +1.7]$ & $-0.9\,[-6.5, +5.0]$ & $-0.9\,[-6.5, +5.0]$ & $+10.1\,[+6.6, +13.6]$ \\
 & WMEM on -- off & $-3.0\,[-9.9, +1.7]$ & $-0.9\,[-5.1, +3.0]$ & $-0.9\,[-5.1, +3.0]$ & $-2.6\,[-4.3, -0.8]$ \\
 & WMOD on -- off & $-0.2\,[-3.7, +2.8]$ & $-3.4\,[-7.7, +0.5]$ & $-3.4\,[-7.7, +0.5]$ & $+1.3\,[-0.1, +2.8]$ \\
\addlinespace[2pt]
\textbf{GLM-5.2} & B2 -- B1 & $-2.9\,[-8.3, +2.3]$ & $-10.9\,[-18.0, -3.1]$ & $-11.4\,[-18.8, -3.8]$ & $+10.2\,[+7.3, +13.1]$ \\
 & B3 -- B1 & $+1.9\,[-4.7, +8.3]$ & $-3.7\,[-11.2, +3.5]$ & $-3.7\,[-12.6, +3.1]$ & $+15.4\,[+13.1, +17.6]$ \\
 & B3 -- B2 & $+4.9\,[-2.0, +11.7]$ & $+7.1\,[-1.0, +14.3]$ & $+7.7\,[-1.3, +15.2]$ & $+5.1\,[+3.0, +7.2]$ \\
 & WMEM on -- off & $-50.1\,[-56.9, -43.5]$ & $-0.9\,[-7.5, +5.5]$ & $-1.3\,[-10.5, +3.9]$ & $-5.8\,[-8.0, -3.9]$ \\
 & WMOD on -- off & $-5.9\,[-10.8, -1.2]$ & $-13.2\,[-20.2, -5.7]$ & $-12.8\,[-19.0, -3.9]$ & $+9.3\,[+6.9, +11.4]$ \\
\addlinespace[2pt]
\textbf{DeepSeek-V4-Pro} & B2 -- B1 & $-10.2\,[-17.4, -2.8]$ & $-4.3\,[-10.9, +1.9]$ & $-4.3\,[-11.0, +1.9]$ & $+10.1\,[+7.8, +12.2]$ \\
 & B3 -- B1 & $-11.2\,[-18.5, -2.6]$ & $-2.8\,[-8.9, +3.3]$ & $-2.8\,[-8.9, +3.3]$ & $+14.2\,[+11.8, +16.4]$ \\
 & B3 -- B2 & $-1.0\,[-7.9, +5.9]$ & $+1.6\,[-5.4, +9.0]$ & $+1.6\,[-5.4, +9.0]$ & $+4.1\,[+2.3, +6.0]$ \\
 & WMEM on -- off & $-26.7\,[-31.5, -21.8]$ & $+0.2\,[-4.6, +5.1]$ & $+0.2\,[-4.6, +5.1]$ & $-4.5\,[-6.4, -2.8]$ \\
 & WMOD on -- off & $-12.0\,[-17.1, -7.0]$ & $-11.8\,[-18.0, -5.9]$ & $-11.8\,[-18.1, -5.9]$ & $+12.1\,[+10.1, +14.1]$ \\
\addlinespace[2pt]
\textbf{Kimi-K2.6} & B2 -- B1 & $-7.4\,[-14.0, -1.0]$ & $-1.3\,[-6.0, +3.9]$ & $-1.3\,[-6.0, +3.9]$ & $+6.7\,[+4.0, +9.3]$ \\
 & B3 -- B1 & $-25.7\,[-32.7, -18.2]$ & $-2.8\,[-8.6, +3.3]$ & $-2.8\,[-8.6, +3.3]$ & $+12.1\,[+9.7, +14.5]$ \\
 & B3 -- B2 & $-18.3\,[-25.7, -11.2]$ & $-1.4\,[-6.8, +3.9]$ & $-1.4\,[-6.8, +3.9]$ & $+5.4\,[+3.0, +7.8]$ \\
 & WMEM on -- off & $-2.3\,[-7.2, +2.2]$ & $+2.4\,[-2.4, +6.9]$ & $+2.4\,[-2.4, +6.9]$ & $-1.4\,[-3.6, +0.7]$ \\
 & WMOD on -- off & $-10.5\,[-14.5, -6.2]$ & $-4.1\,[-8.2, -0.2]$ & $-4.1\,[-8.2, -0.2]$ & $+2.7\,[+1.0, +4.5]$ \\
\addlinespace[2pt]
\textbf{Gemini-3.1-Pro} & B2 -- B1 & $-3.3\,[-8.8, +1.9]$ & $+1.5\,[-0.8, +4.1]$ & $+1.5\,[-0.8, +4.1]$ & $+10.5\,[+8.1, +12.9]$ \\
 & B3 -- B1 & $-8.0\,[-13.9, -2.3]$ & $-1.8\,[-5.6, +1.6]$ & $-1.8\,[-5.6, +1.6]$ & $+15.9\,[+14.2, +17.7]$ \\
 & B3 -- B2 & $-4.6\,[-10.7, +1.2]$ & $-3.3\,[-6.9, -0.0]$ & $-3.3\,[-6.9, -0.0]$ & $+5.4\,[+3.5, +7.4]$ \\
 & WMEM on -- off & $-0.9\,[-5.3, +3.6]$ & $+1.8\,[-0.5, +4.2]$ & $+1.8\,[-0.5, +4.2]$ & $-0.9\,[-2.1, +0.3]$ \\
 & WMOD on -- off & $-2.4\,[-6.4, +1.6]$ & $+0.7\,[-1.4, +2.8]$ & $+0.7\,[-1.4, +2.8]$ & $-0.5\,[-2.2, +1.2]$ \\
\addlinespace[2pt]
\textbf{GPT-5.6-sol} & B2 -- B1 & $-1.4\,[-3.5, +0.6]$ & $+0.5\,[-1.4, +2.2]$ & $+0.5\,[-1.4, +2.2]$ & $+11.3\,[+9.0, +13.6]$ \\
 & B3 -- B1 & $-1.7\,[-4.4, +0.7]$ & $+0.4\,[-1.4, +2.1]$ & $+0.4\,[-1.4, +2.1]$ & $+14.0\,[+11.8, +16.3]$ \\
 & B3 -- B2 & $-0.3\,[-2.9, +2.2]$ & $-0.1\,[-1.8, +1.7]$ & $-0.1\,[-1.8, +1.7]$ & $+2.7\,[+0.6, +4.7]$ \\
 & WMEM on -- off & $-1.7\,[-3.5, +0.1]$ & $-0.7\,[-2.0, +0.7]$ & $-0.7\,[-2.0, +0.7]$ & $+0.6\,[-0.9, +2.2]$ \\
 & WMOD on -- off & $-1.8\,[-3.6, -0.0]$ & $+0.2\,[-1.2, +1.7]$ & $+0.2\,[-1.2, +1.7]$ & $-4.8\,[-6.5, -3.0]$ \\
\bottomrule
\end{tabular}
\caption{Per-model counterpart to main-text Table 1 for both six-model groups. Each block reports the same five marginal contrasts as point estimates with 95\% hierarchical bootstrap intervals. Models are fixed and are not resampled.}
\label{tab:supp-model-effects-all-models}
\end{table*}

\begin{table*}[p]
\centering
\scriptsize
\setlength{\tabcolsep}{2.0pt}
\renewcommand{\arraystretch}{0.77}
\begin{tabular}{@{}llrrrr@{}}
\toprule
Generator & Conditional contrast & $\Delta G$ & $\Delta C_{\mathrm{obs}}$ & $\Delta C_{\mathrm{adj}}$ & $\Delta R$ \\
\midrule
\multicolumn{6}{@{}l}{\textbf{Qwen series}} \\
\textbf{Qwen2.5-3B} & WMEM at B1 & $+0.4\,[-3.2, +2.7]$ & $-11.9\,[-47.5, +0.0]$ & $-44.4\,[-49.5, +7.5]$ & $-1.1\,[-12.1, +10.9]$ \\
 & WMEM at B2 & $+0.1\,[-4.6, +3.9]$ & $-25.1\,[-70.0, +28.2]$ & $-30.7\,[-58.6, +13.5]$ & $+4.1\,[-11.3, +15.7]$ \\
 & WMEM at B3 & $+1.5\,[-0.1, +3.0]$ & $+9.0\,[-55.1, +64.5]$ & $-0.1\,[-11.9, +9.2]$ & $+0.5\,[-10.7, +15.8]$ \\
 & WMOD at B1 & $-0.4\,[-3.9, +1.4]$ & $-11.9\,[-47.5, +0.0]$ & $-39.4\,[-45.0, +10.0]$ & $-2.8\,[-11.0, +11.8]$ \\
 & WMOD at B2 & $-1.9\,[-6.2, +1.7]$ & $-5.4\,[-38.7, +37.9]$ & $-18.8\,[-44.4, +22.8]$ & $+1.4\,[-9.7, +12.4]$ \\
 & WMOD at B3 & $-0.1\,[-1.5, +1.3]$ & $+13.7\,[-42.1, +74.7]$ & $-1.8\,[-13.0, +6.5]$ & $-13.3\,[-23.2, +0.3]$ \\
\addlinespace[1pt]
\textbf{Qwen2.5-7B} & WMEM at B1 & $-21.4\,[-27.4, -14.5]$ & $+2.9\,[-10.7, +17.2]$ & $+0.6\,[-13.6, +13.2]$ & $-3.1\,[-6.8, +0.4]$ \\
 & WMEM at B2 & $-25.4\,[-31.0, -19.6]$ & $-0.6\,[-13.8, +11.3]$ & $-0.6\,[-15.8, +11.3]$ & $+1.1\,[-2.1, +4.3]$ \\
 & WMEM at B3 & $-20.4\,[-26.8, -13.8]$ & $-0.5\,[-10.0, +9.8]$ & $+2.0\,[-10.0, +10.7]$ & $+0.8\,[-2.2, +3.7]$ \\
 & WMOD at B1 & $-40.4\,[-47.3, -33.6]$ & $+31.7\,[+17.4, +47.7]$ & $-58.6\,[-72.2, -41.3]$ & $-17.4\,[-20.7, -14.2]$ \\
 & WMOD at B2 & $-50.2\,[-56.3, -43.5]$ & $+26.4\,[+17.0, +36.4]$ & $-63.6\,[-75.5, -52.9]$ & $-19.6\,[-22.4, -16.3]$ \\
 & WMOD at B3 & $-43.5\,[-50.4, -36.8]$ & $+19.9\,[+7.5, +31.7]$ & $-72.6\,[-83.9, -58.2]$ & $-19.5\,[-22.8, -16.0]$ \\
\addlinespace[1pt]
\textbf{Qwen2.5-14B} & WMEM at B1 & $-4.9\,[-9.3, -0.5]$ & $-12.1\,[-42.4, +18.9]$ & $+8.6\,[-11.4, +18.6]$ & $-3.9\,[-9.6, +1.8]$ \\
 & WMEM at B2 & $-9.3\,[-13.2, -5.7]$ & $-14.7\,[-41.9, +15.9]$ & $-9.4\,[-19.3, -1.0]$ & $-2.2\,[-7.9, +3.3]$ \\
 & WMEM at B3 & $-7.6\,[-12.3, -3.2]$ & $-23.9\,[-53.3, +3.0]$ & $-10.7\,[-23.5, +2.9]$ & $+0.6\,[-3.4, +4.8]$ \\
 & WMOD at B1 & $-24.5\,[-31.5, -17.3]$ & $+10.7\,[-21.0, +44.3]$ & $-54.9\,[-65.9, -32.1]$ & $-23.2\,[-27.9, -18.5]$ \\
 & WMOD at B2 & $-25.7\,[-31.7, -19.3]$ & $+6.4\,[-19.9, +36.8]$ & $-34.8\,[-43.1, -26.1]$ & $-9.6\,[-15.2, -4.2]$ \\
 & WMOD at B3 & $-25.5\,[-31.4, -19.5]$ & $+27.4\,[-6.2, +51.4]$ & $-34.1\,[-48.5, -23.1]$ & $-3.4\,[-8.1, +1.1]$ \\
\addlinespace[1pt]
\textbf{Qwen2.5-32B} & WMEM at B1 & $-29.8\,[-38.9, -21.0]$ & $-3.3\,[-10.8, +3.6]$ & $-3.3\,[-10.9, +3.6]$ & $-2.3\,[-5.9, +1.3]$ \\
 & WMEM at B2 & $-42.4\,[-50.2, -34.6]$ & $-3.2\,[-14.7, +8.0]$ & $-3.2\,[-14.9, +7.9]$ & $-1.5\,[-4.6, +1.7]$ \\
 & WMEM at B3 & $-42.9\,[-50.4, -35.3]$ & $+0.7\,[-9.1, +10.0]$ & $+0.7\,[-10.5, +9.8]$ & $+0.1\,[-3.7, +3.8]$ \\
 & WMOD at B1 & $-23.8\,[-33.0, -14.5]$ & $+1.2\,[-7.9, +9.9]$ & $+1.2\,[-7.9, +9.9]$ & $-2.7\,[-5.8, +0.3]$ \\
 & WMOD at B2 & $+3.3\,[-3.5, +10.0]$ & $-5.9\,[-16.8, +5.2]$ & $-5.9\,[-16.8, +5.6]$ & $-2.4\,[-5.2, +0.2]$ \\
 & WMOD at B3 & $-0.7\,[-8.9, +7.2]$ & $+0.5\,[-11.8, +13.0]$ & $+0.5\,[-11.5, +14.3]$ & $-1.9\,[-5.0, +1.1]$ \\
\addlinespace[1pt]
\textbf{Qwen2.5-72B} & WMEM at B1 & $-16.0\,[-24.0, -8.6]$ & $-0.9\,[-8.4, +8.3]$ & $-0.9\,[-8.4, +8.3]$ & $+0.2\,[-2.7, +3.0]$ \\
 & WMEM at B2 & $-19.3\,[-26.4, -12.7]$ & $+2.3\,[-6.1, +11.3]$ & $+2.3\,[-6.1, +11.3]$ & $-3.2\,[-6.9, +0.6]$ \\
 & WMEM at B3 & $-20.0\,[-27.4, -12.8]$ & $-4.9\,[-11.8, +2.2]$ & $-4.9\,[-11.8, +2.2]$ & $-0.3\,[-3.1, +2.5]$ \\
 & WMOD at B1 & $-7.0\,[-14.4, -0.4]$ & $-4.3\,[-12.5, +3.5]$ & $-4.3\,[-12.5, +3.5]$ & $-2.3\,[-6.4, +1.5]$ \\
 & WMOD at B2 & $-2.1\,[-8.3, +4.3]$ & $-5.1\,[-11.6, +1.4]$ & $-5.1\,[-11.6, +1.4]$ & $-5.4\,[-8.2, -2.5]$ \\
 & WMOD at B3 & $+4.6\,[-1.0, +10.9]$ & $-4.6\,[-14.3, +4.2]$ & $-4.6\,[-14.3, +4.2]$ & $-3.7\,[-7.0, -0.5]$ \\
\addlinespace[1pt]
\textbf{Qwen3.5-397B-A17B} & WMEM at B1 & $-19.9\,[-26.4, -13.4]$ & $-2.4\,[-9.7, +5.0]$ & $-2.4\,[-9.7, +5.0]$ & $-1.0\,[-3.6, +1.7]$ \\
 & WMEM at B2 & $-22.0\,[-30.1, -14.2]$ & $-8.1\,[-19.6, +2.2]$ & $-8.1\,[-19.6, +2.2]$ & $-7.5\,[-11.2, -4.1]$ \\
 & WMEM at B3 & $-24.1\,[-31.3, -17.0]$ & $-4.2\,[-14.6, +5.6]$ & $-4.2\,[-14.6, +5.6]$ & $-4.7\,[-7.4, -1.8]$ \\
 & WMOD at B1 & $+0.2\,[-4.7, +4.9]$ & $-0.5\,[-7.5, +5.8]$ & $-0.5\,[-7.5, +5.8]$ & $+1.7\,[-1.5, +4.9]$ \\
 & WMOD at B2 & $+3.1\,[-4.1, +10.0]$ & $-6.9\,[-18.5, +4.5]$ & $-6.9\,[-18.5, +4.5]$ & $+6.3\,[+3.9, +8.8]$ \\
 & WMOD at B3 & $-12.6\,[-20.9, -4.2]$ & $-8.7\,[-18.8, +1.3]$ & $-8.7\,[-18.8, +1.3]$ & $+10.7\,[+7.5, +13.7]$ \\
\addlinespace[1pt]
\addlinespace[2pt]
\multicolumn{6}{@{}l}{\textbf{Frontier models}} \\
\textbf{Grok-4.3} & WMEM at B1 & $-3.5\,[-11.4, +0.0]$ & $-2.1\,[-8.2, +4.0]$ & $-2.1\,[-8.2, +4.0]$ & $-1.7\,[-4.5, +1.1]$ \\
 & WMEM at B2 & $-2.5\,[-10.7, +2.9]$ & $+0.5\,[-6.8, +7.2]$ & $+0.5\,[-6.8, +7.2]$ & $-2.4\,[-5.8, +0.8]$ \\
 & WMEM at B3 & $-3.1\,[-10.0, +4.2]$ & $-1.3\,[-9.7, +6.6]$ & $-1.3\,[-9.7, +6.6]$ & $-3.6\,[-6.9, -0.4]$ \\
 & WMOD at B1 & $-1.2\,[-4.8, +0.0]$ & $-7.2\,[-13.4, -1.4]$ & $-7.2\,[-13.4, -1.4]$ & $-2.0\,[-5.0, +1.1]$ \\
 & WMOD at B2 & $-0.1\,[-3.3, +2.9]$ & $+1.4\,[-6.2, +8.9]$ & $+1.4\,[-6.2, +8.9]$ & $+1.5\,[-1.7, +4.6]$ \\
 & WMOD at B3 & $+0.6\,[-4.3, +5.6]$ & $-4.6\,[-12.7, +3.0]$ & $-4.6\,[-12.7, +3.0]$ & $+4.5\,[+1.9, +7.3]$ \\
\addlinespace[1pt]
\textbf{GLM-5.2} & WMEM at B1 & $-53.6\,[-62.1, -44.6]$ & $+5.9\,[-4.7, +16.0]$ & $+5.9\,[-6.9, +15.0]$ & $-7.1\,[-10.6, -4.0]$ \\
 & WMEM at B2 & $-43.8\,[-52.8, -34.3]$ & $-13.1\,[-26.2, +1.7]$ & $-14.3\,[-30.4, -1.1]$ & $-5.1\,[-8.5, -2.1]$ \\
 & WMEM at B3 & $-52.9\,[-61.8, -43.9]$ & $+4.4\,[-7.2, +16.1]$ & $+4.4\,[-11.6, +14.4]$ & $-5.1\,[-8.1, -2.5]$ \\
 & WMOD at B1 & $-12.0\,[-17.7, -6.2]$ & $-13.5\,[-23.7, -3.1]$ & $-13.5\,[-24.0, -3.2]$ & $+8.2\,[+5.0, +11.1]$ \\
 & WMOD at B2 & $-7.0\,[-15.5, +1.3]$ & $-9.9\,[-22.2, +3.9]$ & $-8.7\,[-20.2, +5.8]$ & $+9.6\,[+5.6, +13.4]$ \\
 & WMOD at B3 & $+1.4\,[-5.4, +8.2]$ & $-16.2\,[-26.0, -6.6]$ & $-16.2\,[-24.9, -1.5]$ & $+10.0\,[+6.3, +13.7]$ \\
\addlinespace[1pt]
\textbf{DeepSeek-V4-Pro} & WMEM at B1 & $-25.4\,[-33.3, -17.1]$ & $-2.2\,[-11.0, +6.1]$ & $-2.2\,[-11.0, +6.1]$ & $-2.0\,[-4.4, +0.5]$ \\
 & WMEM at B2 & $-34.3\,[-43.5, -25.5]$ & $+1.3\,[-8.1, +10.7]$ & $+1.3\,[-8.2, +10.7]$ & $-6.9\,[-10.0, -3.9]$ \\
 & WMEM at B3 & $-20.4\,[-30.1, -10.4]$ & $+1.6\,[-7.4, +10.5]$ & $+1.6\,[-7.6, +10.5]$ & $-4.7\,[-8.7, -0.9]$ \\
 & WMOD at B1 & $-7.4\,[-15.9, +0.3]$ & $-14.9\,[-23.9, -6.7]$ & $-14.9\,[-23.9, -6.7]$ & $+8.3\,[+4.8, +11.5]$ \\
 & WMOD at B2 & $-16.2\,[-24.2, -8.6]$ & $-14.7\,[-23.8, -5.4]$ & $-14.7\,[-23.8, -5.4]$ & $+13.3\,[+9.6, +16.8]$ \\
 & WMOD at B3 & $-12.3\,[-21.4, -3.5]$ & $-5.8\,[-16.3, +3.7]$ & $-5.8\,[-16.4, +3.7]$ & $+14.6\,[+10.7, +19.1]$ \\
\addlinespace[1pt]
\textbf{Kimi-K2.6} & WMEM at B1 & $-2.7\,[-9.8, +4.5]$ & $+3.1\,[-3.3, +9.2]$ & $+3.1\,[-3.3, +9.2]$ & $+0.4\,[-2.0, +2.8]$ \\
 & WMEM at B2 & $-9.8\,[-20.6, +0.1]$ & $+1.9\,[-5.2, +8.9]$ & $+1.9\,[-5.2, +8.9]$ & $-4.4\,[-8.9, -0.2]$ \\
 & WMEM at B3 & $+5.6\,[-2.3, +12.9]$ & $+2.0\,[-5.3, +9.6]$ & $+2.0\,[-5.3, +9.6]$ & $-0.1\,[-3.4, +3.2]$ \\
 & WMOD at B1 & $-12.6\,[-20.1, -5.4]$ & $-5.0\,[-11.5, +1.3]$ & $-5.0\,[-11.5, +1.3]$ & $+2.0\,[-0.8, +4.8]$ \\
 & WMOD at B2 & $-8.8\,[-14.6, -2.7]$ & $+2.3\,[-5.1, +9.9]$ & $+2.3\,[-5.1, +9.9]$ & $+3.4\,[+0.1, +6.6]$ \\
 & WMOD at B3 & $-10.1\,[-19.5, -0.7]$ & $-9.6\,[-19.0, -1.3]$ & $-9.6\,[-19.0, -1.3]$ & $+2.7\,[-0.2, +5.6]$ \\
\addlinespace[1pt]
\textbf{Gemini-3.1-Pro} & WMEM at B1 & $-4.6\,[-10.1, +0.8]$ & $+1.0\,[-1.3, +3.6]$ & $+1.0\,[-1.3, +3.6]$ & $+1.0\,[-1.2, +3.2]$ \\
 & WMEM at B2 & $-7.8\,[-16.2, +0.3]$ & $-2.3\,[-4.7, -0.3]$ & $-2.3\,[-4.7, -0.3]$ & $-4.0\,[-6.4, -1.4]$ \\
 & WMEM at B3 & $+9.6\,[-2.1, +20.8]$ & $+6.5\,[+0.5, +13.5]$ & $+6.5\,[+0.5, +13.5]$ & $+0.3\,[-2.0, +2.6]$ \\
 & WMOD at B1 & $-3.6\,[-9.9, +2.3]$ & $-1.8\,[-6.2, +1.7]$ & $-1.8\,[-6.2, +1.7]$ & $-1.5\,[-4.1, +1.1]$ \\
 & WMOD at B2 & $-3.2\,[-10.7, +4.2]$ & $-0.9\,[-3.5, +1.4]$ & $-0.9\,[-3.5, +1.4]$ & $+0.0\,[-3.3, +3.1]$ \\
 & WMOD at B3 & $-0.4\,[-6.6, +6.0]$ & $+4.6\,[-0.1, +10.0]$ & $+4.6\,[-0.1, +10.0]$ & $+0.1\,[-2.5, +2.7]$ \\
\addlinespace[1pt]
\textbf{GPT-5.6-sol} & WMEM at B1 & $-2.7\,[-5.7, +0.0]$ & $-1.0\,[-3.6, +1.7]$ & $-1.0\,[-3.6, +1.7]$ & $+2.1\,[-1.1, +5.5]$ \\
 & WMEM at B2 & $+1.1\,[-2.9, +5.0]$ & $-1.1\,[-3.7, +0.5]$ & $-1.1\,[-3.7, +0.5]$ & $+0.1\,[-2.9, +3.1]$ \\
 & WMEM at B3 & $-3.5\,[-6.7, -0.6]$ & $-0.0\,[-2.2, +2.7]$ & $-0.0\,[-2.2, +2.7]$ & $-0.2\,[-2.5, +2.1]$ \\
 & WMOD at B1 & $-0.8\,[-3.9, +2.2]$ & $-1.0\,[-3.6, +1.6]$ & $-1.0\,[-3.6, +1.6]$ & $-12.3\,[-15.6, -8.9]$ \\
 & WMOD at B2 & $-3.1\,[-7.1, +0.8]$ & $+1.4\,[-0.0, +4.1]$ & $+1.4\,[-0.0, +4.1]$ & $-1.8\,[-5.1, +1.4]$ \\
 & WMOD at B3 & $-1.7\,[-6.0, +2.3]$ & $-0.0\,[-2.2, +2.7]$ & $-0.0\,[-2.2, +2.7]$ & $-0.2\,[-2.9, +2.5]$ \\
\bottomrule
\end{tabular}
\caption{Per-model module effects within each base for both six-model groups. Each module is compared on versus off while averaging over the other. Entries are point estimates with 95\% hierarchical bootstrap intervals. Models are fixed and are not resampled.}
\label{tab:supp-model-conditional-all-models}
\end{table*}

\end{document}